\pdfoutput=1
\documentclass[11pt]{article}

\usepackage[]{packages/acl}

\usepackage{times}
\usepackage{latexsym}
\usepackage[T1]{fontenc} % For proper rendering of Latin characters.
\usepackage[utf8]{inputenc}
\usepackage{microtype} % improve the layout and save some space.

\usepackage[show]{packages/chato-notes} % notes
\usepackage[pdftex]{graphicx}           % graphs
\usepackage{subcaption}                 % subfigures
\usepackage{booktabs}                   % professional-quality tables
\usepackage{multirow}                   % multiple rows
\usepackage{amsmath}                    % mathematics
\usepackage{amsfonts}                   % blackboard math symbols
\usepackage{amssymb}                    % 
\usepackage{bbm}                        %
\usepackage{nicefrac}                   % compact symbols for 1/2, etc.
\usepackage{pifont}                     % cross mark
\usepackage{enumitem}                   % vertical space in lists
\usepackage{cleveref}                   % clever references
\usepackage{orcidlink}                  % for ORCID icon.
\usepackage[table]{xcolor}
\usepackage{colortbl}
\usepackage{eurosym} % euro symbol 
\usepackage{soul}
\usepackage{tablefootnote}
\usepackage{float}
\usepackage{svg}

\definecolor{pastelyellow}{RGB}{253, 253, 150}
\sethlcolor{pastelyellow}
\definecolor{red1}{HTML}{FF9999}
\definecolor{orange1}{HTML}{FFB266}
\definecolor{orange2}{HTML}{FFCC99}
\definecolor{green1}{HTML}{BFDBB2}

\newcommand{\citetmp}{\textcolor{red}{(citation)}}%

\newcommand{\at}{\texttt{@}}%
\newcommand{\cmark}{\ding{51}}%
\newcommand{\xmark}{\ding{55}}%
\newcommand{\sub}[1]{\small$_{\textsc{#1}}$}
\newcommand{\sd}[1]{\scriptsize$\pm #1$}
\newcommand{\better}[1]{\scriptsize($+#1\%$)}
\newcommand{\worse}[1]{\scriptsize($-#1\%$)}
\definecolor{shade}{gray}{0.6}
\newcommand{\shade}[1]{\textcolor{shade}{#1}}
\newcommand{\graymidrule}{\arrayrulecolor{lightgray}\midrule\arrayrulecolor{black}}
\newcommand{\graypartialmidrule}[1]{\arrayrulecolor{lightgray}\cmidrule(lr){#1}\arrayrulecolor{black}}
\newcommand{\overbar}[1]{\mkern 1.5mu\overline{\mkern-1.5mu#1\mkern-1.5mu}\mkern 1.5mu}

\newcommand{\orcid}[1]{\textsuperscript{\normalsize\orcidlink{#1}}}

\setlist[itemize,enumerate]{itemsep=-5pt, topsep=1pt}

\title{Know When to Fuse: Investigating Non-English Hybrid Retrieval\\ in the Legal Domain}
\author{
Antoine Louis\orcid{https://orcid.org/0000-0001-8392-3852}\textnormal{,} 
Gijs van Dijck\orcid{https://orcid.org/0000-0003-4102-4415}\textnormal{,} 
Gerasimos Spanakis\orcid{https://orcid.org/0000-0002-0799-0241} \\
Law \& Tech Lab, Maastricht University \\
{\small \texttt{\{a.louis, gijs.vandijck, jerry.spanakis\}@maastrichtuniversity.nl}} \\
}

\begin{document}
\maketitle

\begin{abstract}
Hybrid search has emerged as an effective strategy to offset the limitations of different matching paradigms, especially in out-of-domain contexts where notable improvements in retrieval quality have been observed. However, existing research predominantly focuses on a limited set of retrieval methods, evaluated in pairs on domain-general datasets exclusively in English. In this work, we study the efficacy of hybrid search across a variety of prominent retrieval models within the unexplored field of law in the French language, assessing both zero-shot and in-domain scenarios. Our findings reveal that in a zero-shot context, fusing different domain-general models consistently enhances performance compared to using a standalone model, regardless of the fusion method. Surprisingly, when models are trained in-domain, we find that fusion generally diminishes performance relative to using the best single system, unless fusing scores with carefully tuned weights. These novel insights, among others, expand the applicability of prior findings across a new field and language, and contribute to a deeper understanding of hybrid search in non-English specialized domains.\footnote{Our source code and models are available at \url{https://github.com/maastrichtlawtech/fusion}.}
\end{abstract}

\begin{figure}[t]
    \centering
    \includegraphics[width=1\columnwidth]{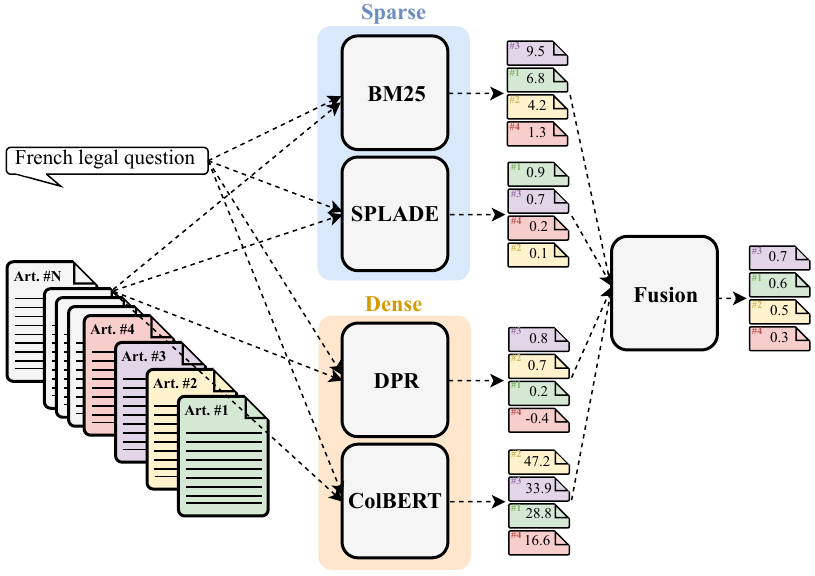}
    \caption{A high-level illustration of the hybrid search workflow based on various sparse and dense retrievers.}
    \label{fig:main}
\end{figure}

%-------------------------------------------------------------------------
%                             INTRODUCTION
%-------------------------------------------------------------------------
\section{Introduction \label{sec:introduction}}
Information retrieval is typically addressed through one of two fundamental matching paradigms: (i) \textsl{lexical matching}, which relies on an exact match of terms between queries and documents; and (ii) \textsl{semantic matching}, which measures complex relationships between words to capture underlying semantics. Lexical matching is simple, efficient, and generally effective across various domains \citep{thakur2021beir}. However, it suffers from the vocabulary gap issue \citep{berger2000bridging}, where relevant information might not explicitly include query terms yet still fulfills the actual informational needs. Semantic models remedy vocabulary mismatches by learning to model semantic similarity, resulting in significant in-domain performance gains \citep{qu2021rocketqa, xiong2021approximate, hofstatter2021efficiently}. Nevertheless, these models tend to exhibit limited generalization across unseen topics \citep{thakur2021beir}, which is particularly problematic in highly specialized domains, like law, where high-quality labeled data is both scarce and costly.
%these models can still struggle with simple factual questions \citep{sciavolino2021simple}

Recent works suggest that combining these two paradigms can enhance retrieval quality \citep{kuzi2020leveraging, wang2021bert, ma2021replication}, particularly in out-of-distribution settings \citep{chen2022out, bruch2024analysis}, as they tend to mitigate each other's limitations. However, these efforts have mostly been limited to combining no more than two systems -- typically pairing BM25 \citep{robertson1994okapi} with single-vector dense bi-encoders \citep{reimers2019sentence} -- while constraining evaluation to English datasets only.
%without investigating variations within similar matching paradigms.

Our work aims to extend this scope by investigating the potential synergies among a broader range of retrieval models, encompassing both sparse and dense methods, specifically within the uncharted \textsl{legal} domain in the \textsl{French} language, as illustrated in \Cref{fig:main}. Our contributions are threefold:
\begin{itemize}
    \item First, we investigate the efficacy of combining diverse domain-general retrieval models for legal retrieval, assuming \textsl{no} domain-specific labeled data is available -- a highly usual situation in specialized domains like law.
    \item Second, we explore the extent to which specialized retrievers and their fusion can impact in-domain performance, assuming \textsl{limited} domain-specific training data is available.
    \item Finally, we release all our learned retrievers, including the first French SPLADE and ColBERT models for general and legal domains.
\end{itemize}

%-------------------------------------------------------------------------
%                       METHODOLOGY
%-------------------------------------------------------------------------
\section{Methodology\label{sec:methodology}}
Assuming that different matching paradigms may be complementary in how they model relevance \citep{chen2022out, bruch2024analysis}, we aim to explore the potential of combining various systems to enhance performance on French legal retrieval. In this section, we outline the retrieval models (\S\ref{subsec:retrieval_models}), fusion techniques (\S\ref{subsec:fusion_techniques}), and experimental setup (\S\ref{subsec:experimental_setup}) employed in our study, with additional comprehensive details available in Appendix \ref{app:methodology_details}.

\subsection{Retrieval Models\label{subsec:retrieval_models}}
We select several prominent retrieval methods representing diverse matching paradigms, all demonstrating high effectiveness in prior studies. Specifically, we explore the unsupervised BM25 weighting scheme \citep{robertson1994okapi}, our own \textsl{single-vector dense} \citep{lee2019latent,chang2020pretraining,karpukhin2020dense}, \textsl{multi-vector dense} \citep{khattab2020colbert, santhanam2022colbertv2}, and \textsl{single-vector sparse} \citep{formal2021spladev2, formal2021splade} bi-encoder models -- respectively dubbed DPR{\sub{fr}}, ColBERT{\sub{fr}}, and SPLADE{\sub{fr}} -- and a \textsl{cross-attention} model \citep{nogueira2019passage, han2020learning, gao2021rethink} termed monoBERT{\sub{fr}}. Following a preliminary comparative analysis of various pretrained French language models in \Cref{app:backbone}, we choose CamemBERT{\sub{base}} \citep{martin2020camembert} as the backbone encoder for all our supervised neural retrievers. We refer readers to \Cref{app:retrieval_models} for detailed explanations of each method's relevance matching and optimization processes.

\subsection{Fusion Techniques\label{subsec:fusion_techniques}}
To leverage existing retrieval methods without modification, our study explores \textsl{late} fusion techniques, which aggregate results post-prediction -- in contrast to early fusion methods that merge latent representations of distinct retrievers within the feature space prior to making predictions. In this context, the relevance of a candidate can be assessed using two main measures: its position in the ranked list or its predicted score. This distinction underpins the two primary late fusion approaches explored in this study: \textsl{score-based} and \textsl{rank-based} fusion. Specifically, we investigate \textsl{normalized score fusion} (NSF; \citealp{lee1995combining}) with various scaling techniques, \textsl{Borda count fusion} (BCF; \citealp{ho1994decision}), and \textsl{reciprocal rank fusion} (RRF; \citealp{cormack2009reciprocal}). See \Cref{app:fusion_techniques} for detailed definitions of each method.

\setlength{\tabcolsep}{10pt}
\begin{table*}[t]
\centering
\resizebox{\textwidth}{!}{%
\begin{tabular}{ll|cc|cc|cc|cc|cc}
\toprule
& \multirow{2}{*}{\textbf{Model}} & \multicolumn{2}{c|}{\textbf{mMARCO-fr}} & \multicolumn{2}{c|}{\textbf{Model Size}} & \multicolumn{2}{c|}{\textbf{\#Samples}} & \multicolumn{2}{c|}{\textbf{Batch Size}} & \multicolumn{2}{c}{\textbf{Hardware}} \\ 
&& MRR\at10 & R\at500 & \#Params & RAM & PF & F & PF & F & Pre-Finetune & Finetune \\
\midrule
\multicolumn{12}{l}{\textbf{Baselines}} \\
\shade{1} & BM25 {\footnotesize$(k1\!=\!0.9, b\!=\!0.4)$} & 0.143 & 0.681 &     -- &    -- &   -- &   -- &  -- & --   & --             & --            \\
\shade{2} & mE5{\sub{small}}        & 0.297 & 0.908 & 117.7M & 0.5GB &   1B & 1.6M & 32k & 512  & 32$\times$V100 & 8$\times$V100 \\
\shade{3} & mE5{\sub{base}}         & 0.303 & \underline{0.914} & 278.0M & 1.1GB &   1B & 1.6M & 32k & 512  & 64$\times$V100 & 8$\times$V100 \\
\shade{4} & mE5{\sub{large}}        & \underline{0.311} & 0.909 & 559.9M & 2.2GB &   1B & 1.6M & 32k & 512  & \textsl{Unk.}  & 8$\times$V100 \\
\shade{5} & BGE-M3{\sub{dense}}     & 0.270 & 0.891 & 567.8M & 2.3GB & 1.2B & 1.6M & 67k & 1.2k & 96$\times$A800 & 24$\times$A800 \\
\midrule
\multicolumn{12}{l}{\textbf{Learned models (ours)}} \\
\shade{6} & DPR{\sub{fr-base}}      & 0.285 & 0.891 & 110.6M & 0.4GB &   -- & 0.5M & --   & 152  & --            & 1$\times$V100 \\
\shade{7} & SPLADE{\sub{fr-base}}   & 0.247 & 0.860 & 110.6M & 0.4GB &   -- & 0.5M & --   & 128  & --            & 1$\times$H100 \\
\shade{8} & ColBERT{\sub{fr-base}}  & \hspace{5pt}0.295$^\dag$ & \hspace{5pt}0.884$^\dag$ & 110.6M & 0.4GB &   -- & 0.5M & --   & 128  & --            & 1$\times$H100 \\
\shade{9} & monoBERT{\sub{fr-base}} & \hspace{5pt}\textbf{0.334}$^\star$ & \hspace{5pt}\textbf{0.965}$^\star$ & 110.6M & 0.4GB &   -- & 0.5M & --   & 128  & --            & 1$\times$H100 \\
\bottomrule
\multicolumn{12}{l}{\footnotesize$^\dag$ Evaluated using the PLAID retrieval engine \citep{santhanam2022plaid}.\quad $^\star$ Evaluated by re-ranking 1k candidates including gold and hard negative passages.} \\
\end{tabular}
}
\caption{Retrieval results on mMARCO-fr small dev set (in-domain). We report each model's training resources.}
\label{tab:mmarco_results}
\end{table*}
\setlength{\tabcolsep}{6pt}

\subsection{Experimental Setup \label{subsec:experimental_setup}}

\paragraph{Datasets.}
We exploit two French text ranking datasets: the domain-general mMARCO-fr \citep{bonifacio2021mmarco} and the domain-specific LLeQA \citep{louis2024lleqa}. The former is a translated version of MS MARCO \citep{nguyen2018msmarco} in 13 languages, including French. It comprises a corpus of 8.8M passages, 539K training queries, and 6980 development queries. LLeQA targets long-form question answering and information retrieval within the legal domain. It consists of 1,868 French-native questions on various legal topics, distributed across training (1472), development (201), and test (195) sets. Each question is expertly annotated with references to relevant legal provisions drawn from a corpus of 27,942 Belgian law articles.
%, whose average length is 157 words.
% which are an order of magnitude longer than the questions, as shown in \Cref{fig:lleqa_seq_lengths}.

\paragraph{Evaluation metrics.}
To measure effectiveness, we use official metrics for each dataset: mean reciprocal rank at cutoff 10 (MRR\at10) for mMARCO, and average r-precision (RP) for LLeQA. Both metrics are rank-aware, meaning they are sensitive to variations in the ordering of retrieved results. Additionally, we report the rank-unaware recall measure at various cutoffs (R\at$k$), which is particularly useful for assessing performance of first-stage retrievers. See \Cref{app:evaluation_metrics} for details.

% \begin{figure}[t]
%     \centering
%     \includegraphics[width=1\linewidth]{img/length_distributions_proportion.pdf}
%     \caption{Distributions of tokenized sequence lengths for questions and articles in LLeQA.}
%     \label{fig:lleqa_seq_lengths}
% \end{figure}
% The $x$-axis is in log scale.
% This figure motivates our choice of chunking queries at 64 and articles at 512 when training on LLeQA.

\paragraph{Baselines.}
We evaluate our learned retrievers and their hybrid configurations against leading open-source multilingual retrieval models, including BM25 \citep{robertson1994okapi}, mE5 \citep{wang2024me5} in its small, base, and large variants, and BGE-M3 \citep{chen2024bge} in its dense version.

\subsection{Efficiency\label{subsec:efficiency}}
To evaluate the practicality of each system for real-world deployment, we assess their computational and memory efficiency during inference.

\paragraph{Index size.}
We start by calculating the storage footprint of the indexed LLeQA articles, pre-computed offline and loaded at inference, noting that the indexing method varies with the retrieval approach. Sparse methods like BM25 and SPLADE use inverted indexes, which store each vocabulary term along lists of articles containing the term and its frequency within those articles. Single-vector dense models, such as DPR{\sub{fr}}, mE5, and BGE-M3, rely on flat indexes for brute-force search, sequentially storing vectors on $d \times b \times |\mathcal{C}|$ bits given $d$-dimensional representations of articles from corpus $\mathcal{C}$ encoded in $b$ bits (with $b\!=\!32$ in our study).\footnote{While ANNS indexes such as HNSW \citep{malkov2014approximate} enable more efficient retrieval, they introduce significant overhead which make flat indexes generally preferable for smaller datasets like LLeQA \citep{milvus2022flat, redis2024vectors}.} Meanwhile, ColBERT uses an advanced centroid-based indexing to store late-interaction token embeddings, with a footprint comparable to dense flat indexes \citep{santhanam2022colbertv2}.
%\citep{redis2024vectors, weaviate2024flat, pinecone2024nearest, milvus2022flat}
%\footnote{While approximate nearest neighbor search indexes such as IVF \citep{sivic2003video} and HNSW \citep{malkov2014approximate} enable more efficient retrieval, they introduce significant overhead which make flat indexes generally preferable for smaller datasets like LLeQA \citep{milvus2022flat, redis2024vectors}.}

\paragraph{Retrieval latency.}
We then measure the retrieval latency per query in seconds. We use a query batch size of one to simulate streaming queries and compute the average latency across all queries in the LLeQA dev set. Measurements are conducted on a single NVIDIA H100 for GPU search and on a AMD EPYC 7763 for CPU search.

\paragraph{Inference FLOPs.}
Finally, we estimate the number of floating point operations (FLOPs) per query as a hardware-agnostic measure of compute usage. Details of our estimation methodology across the different systems are provided in \Cref{app:counting_flops}.

\setlength{\tabcolsep}{10pt}
\begin{table*}[t]
\centering
\resizebox{0.89\textwidth}{!}{%
\begin{tabular}{ll|ccc|rc|cc|l} 
\toprule
& \multirow{2}{*}{\textbf{Model}} & \multicolumn{3}{c|}{\textbf{LLeQA}} & \multicolumn{2}{c|}{\textbf{Index Storage}} & \multicolumn{2}{c|}{\textbf{Latency (s/q)}} & \multirow{2}{*}{\textbf{FLOPs}} \\
&& RP & R\at10 & R\at500 & \multicolumn{1}{c}{Disk{\footnotesize\textsuperscript{\ding{70}}}} & Ratio{\footnotesize\textsuperscript{\ding{168}}} & GPU & CPU & \\
\midrule
\multicolumn{10}{l}{\textbf{Baselines}} \\
\shade{1} & BM25 {\footnotesize$(k1\!=\!2.5, b\!=\!0.2)$} & \textbf{0.163} & 0.367 & 0.672 &   6.6MB\hspace{5pt} & $\times 0.2$ & --     & 0.142 & 1.7e+6 \\
\shade{2} & mE5{\sub{small}}   & 0.081 & 0.174 & 0.611 &  40.9MB\hspace{5pt} & $\times 1.5$ & 0.013 & 0.028 & 6.6e+8 \\
\shade{3} & mE5{\sub{base}}    & 0.074 & 0.157 & 0.653 &  81.9MB\hspace{5pt} & $\times 2.9$ & 0.014 & 0.065 & 2.6e+9 \\
\shade{4} & mE5{\sub{large}}   & 0.074 & 0.194 & 0.695 & 109.1MB\hspace{5pt} & $\times 3.9$ & 0.022 & 0.121 & 9.2e+9 \\
\shade{5} & BGE-M3{\sub{dense}}& 0.090 & 0.325 & 0.734 & 109.1MB\hspace{5pt} & $\times 3.9$ & 0.023 & 0.113 & 9.2e+9 \\
\graymidrule
\multicolumn{10}{l}{\textbf{Learned models (ours)}} \\
\shade{6} & DPR{\sub{fr-base}}      & 0.046 & 0.146 & 0.590 & 81.9MB\hspace{5pt} & $\times 2.9$ & 0.013 & 0.057 & 2.6e+9 \\
\shade{7} & SPLADE{\sub{fr-base}}   & 0.045 & 0.107 & 0.596 & 30.2MB\hspace{5pt} & $\times 1.1$ & 0.013 & 0.609 & 2.6e+9 \\ 
\shade{8} & ColBERT{\sub{fr-base}}  & \hspace{5pt}0.047$^\dag$ & \hspace{5pt}0.148$^\dag$ & \hspace{5pt}0.517$^\dag$ & 185.8MB$^\dag$ & $\times 6.7$ & \hspace{5pt}0.031$^\dag$ & \hspace{5pt}0.142$^\dag$ & 2.6e+11 \\ 
\shade{9} & monoBERT{\sub{fr-base}} & 0.102 & 0.290 & 0.536 & \multicolumn{1}{c}{--} & -- & \hspace{5pt}4.472$^\star$ & \hspace{5pt}184.7$^\star$ & 2.2e+13$^\star$ \\ 
\graymidrule
\multicolumn{10}{l}{\textbf{Hybrid combinations}} \\
\shade{10} & NSF{\sub{z-score}}$(\shade{1},\shade{7})$ & 0.130 & 0.372 & \textbf{0.755} & 36.8MB\hspace{5pt} & $\times 1.3$ & -- & -- & 2.6e+9 \\
\shade{11} & NSF{\sub{min-max}}$(\shade{1},\shade{8})$ & 0.134 & 0.397 & 0.746 & 192.4MB\hspace{5pt} & $\times 6.9$ & -- & -- & 2.6e+11 \\
\shade{12} & NSF{\sub{z-score}}$(\shade{1},\shade{6},\shade{7})$ & 0.092 & 0.354 & 0.742 & 118.7MB\hspace{5pt} & $\times 4.3$ & -- & -- & 5.2e+9 \\
\shade{13} & NSF{\sub{z-score}}$(\shade{1},\shade{7},\shade{8})$ & 0.109 & \underline{0.399} & \underline{0.753} & 222.6MB\hspace{5pt} & $\times 8.0$ & -- & -- & 5.2e+9 \\
\shade{14} & NSF{\sub{z-score}}$(\shade{1},\shade{6},\shade{8})$ & \underline{0.139} & \textbf{0.407} & 0.750 & 274.3MB\hspace{5pt} & $\times 9.8$ & -- & -- & 2.6e+11 \\
\shade{15} & NSF{\sub{z-score}}$(\shade{1},\shade{6},\shade{7},\shade{8})$ & 0.125 & 0.388 & 0.736 & 304.5MB\hspace{5pt} & $\times 10.9$ & -- & -- & 2.7e+11 \\
\bottomrule
\multicolumn{10}{l}{\footnotesize \textsuperscript{\ding{70}} Estimated with 32-bit precision for dense vectors. \quad \textsuperscript{\ding{168}} Ratio of index size to plain text size.} \\
\end{tabular}
}
\caption{Retrieval results on LLeQA test set (zero-shot). We report performance of the best hybrid configurations obtained after extensive evaluation on LLeQA dev set (see \Cref{tab:outdomain_fusion}).}
\label{tab:zeroshot_main}
\end{table*}
\setlength{\tabcolsep}{6pt}
\section{Zero-Shot Evaluation \label{sec:zeroshot_eval}}
In this section, we investigate the out-of-domain generalization capabilities of modern retrieval models trained on a budget and explore the efficacy of their fusion in the specialized domain of law. Specifically, we explore the following question: \textsl{Assuming a lack of domain-specific labeled data and limited computational resources, how effectively can hybrid combinations of domain-general retrieval models perform within the legal domain?} To address this, we train the supervised retrieval models presented in \Cref{sec:methodology} on the French segment of the domain-general mMARCO dataset. We denote the resulting models with the \textsc{fr-base} subscript throughout the rest of the paper.

\paragraph{Main results.}
When evaluated on mMARCO-fr, our learned French retrievers exhibit competitive, and at times superior, in-domain performance compared to leading multilingual retrieval models. This is particularly notable given their relatively smaller size and the constrained resources used during training, as shown in \Cref{tab:mmarco_results}. For instance, DPR{\sub{fr-base}} surpasses BGE-M3{\sub{dense}} with only one-fifth of its parameters, 2400$\times$ fewer training samples, and significantly less training compute. Additionally, our cross-encoder consistently outperforms all other retrieval methods, corroborating prior findings on the efficacy of cross-attention \citep{hofstatter2020improving}.

\noindent However, results in \Cref{tab:zeroshot_main} reveal that, when evaluated in the legal domain, our domain-general French retrievers generally underperform against the multilingual baselines, except for our cross-encoder which remains competitive at smaller cutoffs. This discrepancy is largely due to the baselines' extensive (pre-)finetuning across diverse data with large batch sizes -- which proved beneficial for enhanced contrastive learning \citep{qu2021rocketqa}. Surprisingly, BM25 outperforms all neural models in this specialized context, reaffirming its robustness when dealing with out-of-distribution data.

\noindent Besides, BM25 is notably efficient at inference, with an index up to 30$\times$ smaller and significantly fewer FLOPs than neural retrievers. In contrast, the full interaction mechanism of monoBERT{\sub{fr-base}} incurs substantial computational costs, resulting in latencies up to 350$\times$ and 2350$\times$ higher on GPU and CPU, respectively, than the other learned French models -- while assessed to re-rank 1,000 candidates only rather than the whole corpus. ColBERT{\sub{fr-base}}, with its token-to-token interaction, achieves reasonable latencies on both GPU and CPU due to the low-level optimization of PLAID, but results in a larger index. Meanwhile, SPLADE{\sub{fr-base}} stands out among neural methods by using an inverted index nearly 3$\times$ smaller than that of its single-vector dense counterpart.

\noindent Finally, we observe that fusing BM25 with one or more of our learned domain-general French models consistently and significantly outperforms all individual retrievers in the zero-shot setting (except on RP where BM25 excels) yet at the expense of increased memory -- but comparable latencies when using parallelization. This fusion markedly enhances recall at large cutoffs compared to standalone BM25. On recall\at10, most fusions improve upon BM25; notably, the BM25+DPR{\sub{fr-base}}+ColBERT{\sub{fr-base}} fusion shows a 4\% enhancement and surpasses both DPR{\sub{fr-base}} and ColBERT{\sub{fr-base}} by around 26\%. Surprisingly, the BM25+SPLADE{\sub{fr-base}} fusion is the most effective on R\at500 while standing out for its efficiency due to both methods’ use of inverted indexes.

\begin{figure*}[t]
\centering
\begin{subfigure}[t]{.32\textwidth}
  \centering
  \includegraphics[width=\linewidth]{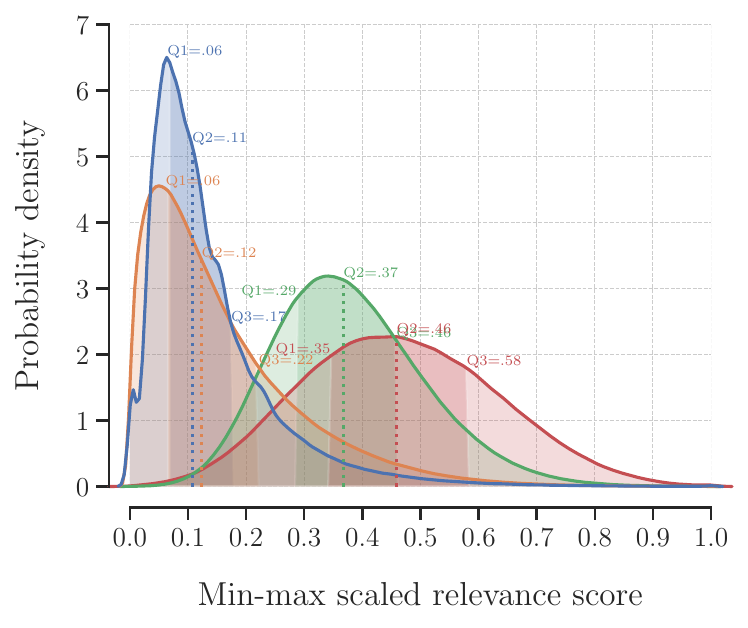}
\end{subfigure}
\begin{subfigure}[t]{.32\textwidth}
  \centering
  \includegraphics[width=\linewidth]{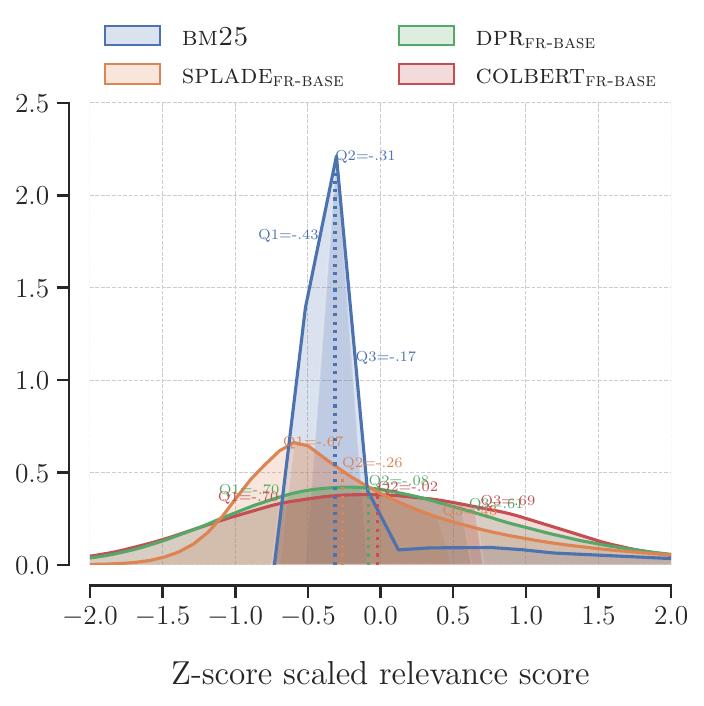}
\end{subfigure}
\begin{subfigure}[t]{.32\textwidth}
  \centering
  \includegraphics[width=\linewidth]{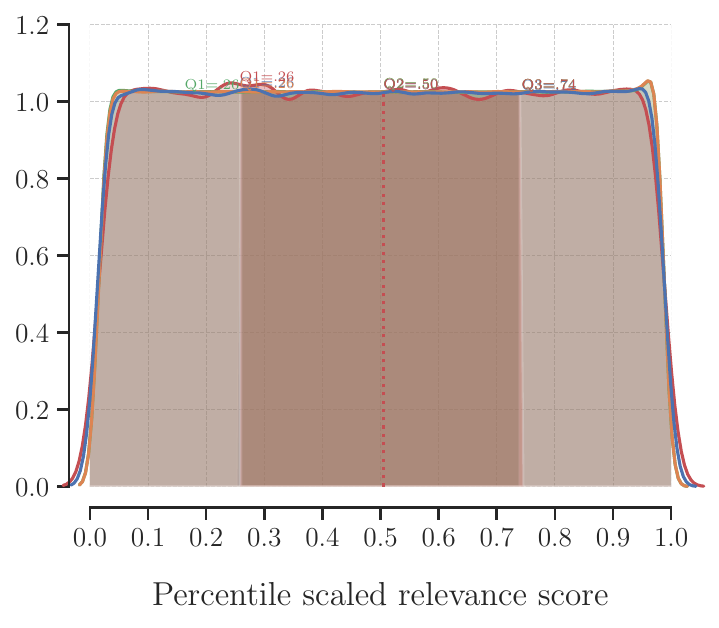}
\end{subfigure}
\caption{In-domain score distributions of domain-general end-to-end retrievers, normalized using min-max, z-score, and percentile scaling. The distributions are derived from ranking all 27,942 articles in LLeQA's knowledge corpus against the 201 development set queries, resulting in approximately 5.6 million scores per system.}
\label{fig:normalized_distributions}
\end{figure*}
%NB: Distributions can only be estimated using the dev set as we later use the training set for finetuning so the in-domain distribution would be biased if we used the training set. Besides, 5.6M data points seems representative enough.

\begin{figure*}[t]
\centering
\begin{subfigure}[t]{.40\textwidth}
  \centering
  \includegraphics[width=\linewidth]{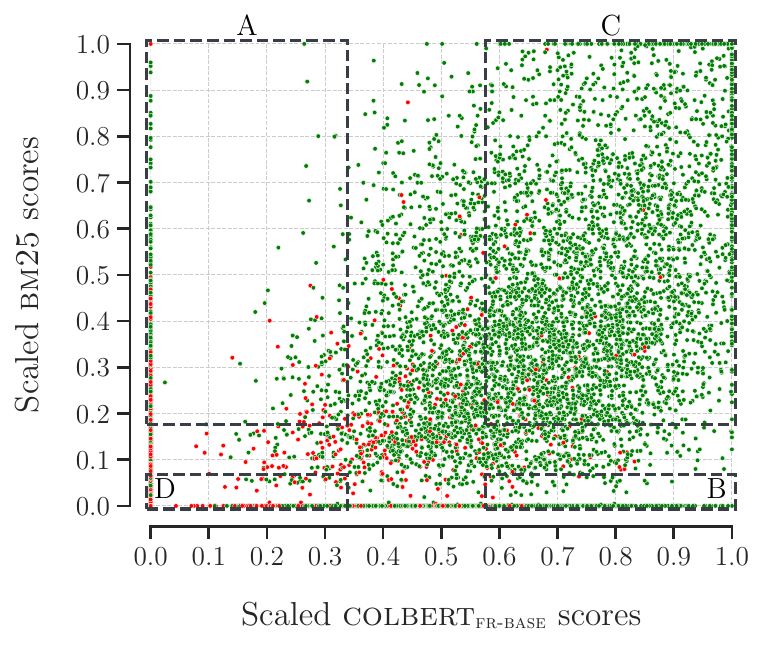}
\end{subfigure}
\hfill
\begin{subfigure}[t]{.55\textwidth}
  \centering
  \includegraphics[width=\linewidth]{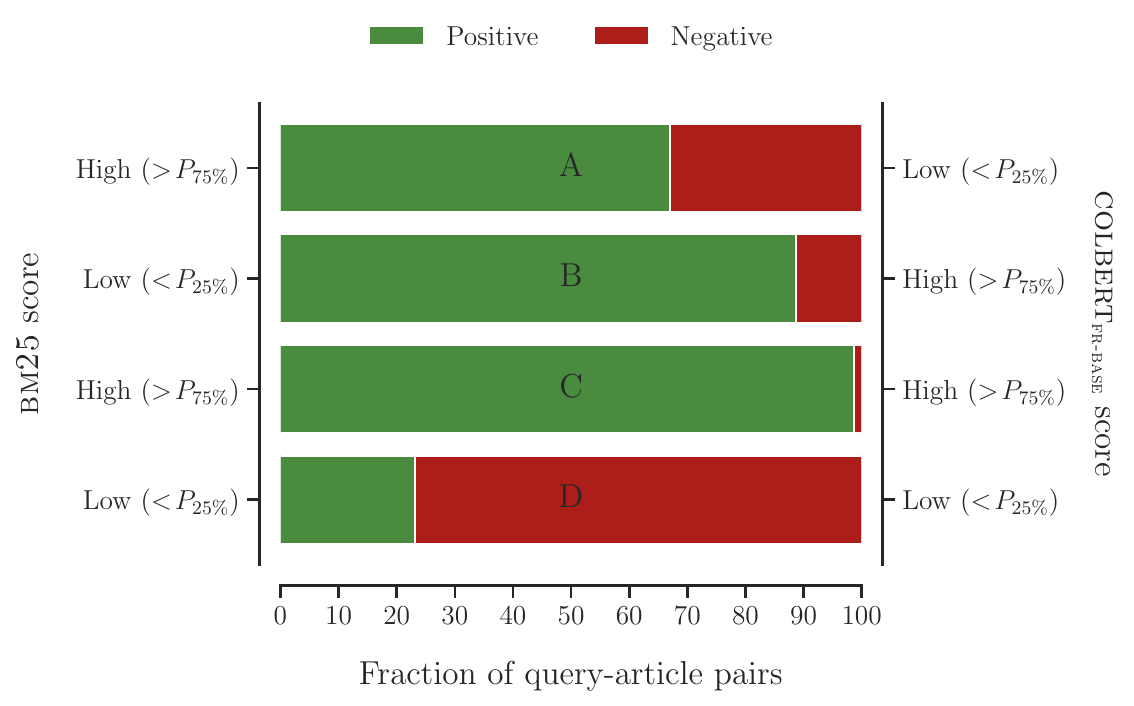}
\end{subfigure}
\caption{Illustration of the complementary relationship between a sparse (BM25) and a dense (ColBERT{\sub{fr-base}}) system on out-of-distribution data. Scores have been min-max normalized and categorized into four distinct regions based on each system's global distribution, depicted in \Cref{fig:normalized_distributions}.}
\label{fig:complementarity}
\end{figure*}
%%While regions where both systems score high (resp., low) predominantly correspond to positive (resp., negative) examples, as seen in C (resp., D), each system effectively compensates for the other's underperformance in identifying positive examples, as observed in A and B.
%¨Displayed in green are 9,330 relevant query-article pairs from LLeQA's training set and in red, an equivalent number of randomly sampled negative pairs.
%, with ``low'' values corresponding to scores below the first quartile and ``high'' ones above the third quartile.

\paragraph{How do score distributions vary across models?}
\Cref{fig:normalized_distributions} depicts the score distributions of end-to-end retrievers, normalized using both traditional techniques and our proposed percentile normalization. We find that traditional scaling methods lead to misaligned distributions among retrievers, particularly under min-max scaling. Such misalignment impacts score fusion as identical scores may convey different levels of relevance across systems. For example, a min-max normalized score of 0.35 approximates the median for DPR{\sub{fr-base}}, but corresponds to the 95th percentile for BM25. When these scores are equally combined, the higher relevance indicated by BM25's score is therefore negated. To address this, we explore a new scaling approach that maps scores to their respective percentiles within each system's overall score distribution, estimated using around 5.6 million data points per system. This way, a score of 0.35 is adjusted to 0.5 for DPR{\sub{fr-base}} and 0.95 for BM25, leading to a relatively higher fused score that favor high relevance signals. This method requires pre-computing each retriever's score distribution, ideally with a volume matching the corpus size to avoid score collisions. Despite its intuitive appeal, our empirical findings reveal that this percentile-based scaling does not surpass traditional methods, as shown in \Cref{tab:outdomain_fusion}. 
% Highlight the issue with unaligned distributions: a score of 0.4 has not the same meaning for BM25 and ColBERT. For the former it is extremely uncommon and associated with the top X% scores (so intuitively highly relevant) while for the latter it fits around the median so at first glance not that relevant as many pairs score higher. This is the motivation for our proposed percentile score fusion. 
%By converting scores to percentiles, you map each score to its position relative to the entire distribution of scores from a given retrieval system. This method inherently adjusts for differences in score distributions across different search models.
%Despite negative results, we believe the rationale behind our approach merits discussion.
%Percentile transformation: if the number of data points (N) in the estimated distribution is too small (e.g., 100) then many documents will have the same score -> increase N. Should N be the corpus size? Or something arbitrary like 1k, 10k or 100k?
%http://www.philender.com/courses/intro/percentile.html
%https://www.education.pa.gov/Documents/K-12/Assessment%20and%20Accountability/PVAAS/Methodology/MakingSenseOfNCEsAndStandardErrors.pdf

\setlength{\tabcolsep}{7pt}
\begin{table*}[t]
\centering
\resizebox{\textwidth}{!}{%
\begin{tabular}{ll|c|c|ll|ll|ll} 
\toprule
&\multirow{2}{*}{\textbf{Method}} & \multirow{2}{*}{\textbf{BCF}} & \multirow{2}{*}{\textbf{RRF}} & \multicolumn{2}{c|}{\textbf{NSF{\sub{min-max}}}} & \multicolumn{2}{c|}{\textbf{NSF{\sub{z-score}}}} & \multicolumn{2}{c}{\textbf{NSF{\sub{percentile}}}} \\
&&&& Equal & Tuned & Equal & Tuned & Equal & Tuned \\
\midrule\midrule
\multirow{4}{*}{\rotatebox[origin=c]{90}{\parbox{52pt}{\centering\footnotesize Single\\baselines}}} & BM25                    & 0.232 & 0.232 & 0.232 & 0.232 & 0.232 & 0.232 & 0.232 & 0.232 \\
& DPR{\sub{fr-base}}      & 0.184 & 0.184 & 0.184 & 0.184 & 0.184 & 0.184 & 0.184 & 0.184 \\
& SPLADE{\sub{fr-base}}   & 0.180 & 0.180 & 0.180 & 0.180 & 0.180 & 0.180 & 0.180 & 0.180 \\
& ColBERT{\sub{fr-base}}  & 0.232 & 0.232 & 0.232 & 0.232 & 0.232 & 0.232 & 0.232 & 0.232 \\
\midrule
\multirow{2}{*}{\rotatebox[origin=c]{90}{\parbox{26pt}{\footnotesize Sparse\\/ dense}}} & BM25 + SPLADE{\sub{fr-base}} & \cellcolor{green1!55}0.262  & \cellcolor{green1!55}0.279  & \cellcolor{green1!55}0.295 & \cellcolor{green1!55}0.295 & \cellcolor{green1!55}0.286 & \cellcolor{green1!55}0.300$^{\dagger}$ & \cellcolor{green1!55}0.282 & \cellcolor{green1!55}0.286 \\
& DPR{\sub{fr-base}} + ColBERT{\sub{fr-base}} & \cellcolor{red1!30}0.219 & \cellcolor{red1!30}0.230 & \cellcolor{red1!30}0.229 & \cellcolor{green1!55}0.243 & \cellcolor{red1!30}0.227 & \cellcolor{green1!55}0.243 & \cellcolor{red1!30}0.206 & \cellcolor{red1!30}0.228 \\
\graymidrule
\multirow{4}{*}{\rotatebox[origin=c]{90}{\parbox{52pt}{\centering\footnotesize Dense$+$sparse\\w. 2 systems}}} & BM25 + DPR{\sub{fr-base}} & \cellcolor{green1!55}0.233  & \cellcolor{green1!55}0.262  & \cellcolor{green1!55}0.268 & \cellcolor{green1!55}0.276 & \cellcolor{green1!55}0.265 & \cellcolor{green1!55}0.286 & \cellcolor{green1!55}0.257 & \cellcolor{green1!55}0.257 \\
& BM25 + ColBERT{\sub{fr-base}} & \cellcolor{green1!55}0.249  & \cellcolor{green1!55}0.269  & \cellcolor{green1!55}0.293 & \cellcolor{green1!55}0.303$^{\dagger}$ & \cellcolor{green1!55}0.262 & \cellcolor{green1!55}0.294 & \cellcolor{green1!55}0.261 & \cellcolor{green1!55}0.266 \\
& SPLADE{\sub{fr-base}} + DPR{\sub{fr-base}} & \cellcolor{green1!55}0.188  & \cellcolor{green1!55}0.203  & \cellcolor{green1!55}0.196 & \cellcolor{green1!55}0.217 & \cellcolor{green1!55}0.197 & \cellcolor{green1!55}0.218 & \cellcolor{green1!55}0.195 & \cellcolor{green1!55}0.210 \\
& SPLADE{\sub{fr-base}} + ColBERT{\sub{fr-base}} & \cellcolor{green1!55}0.238  & \cellcolor{red1!30}0.220 & \cellcolor{red1!30}0.225 & \cellcolor{green1!55}0.249 & \cellcolor{red1!30}0.229 & \cellcolor{green1!55}0.243 & \cellcolor{red1!30}0.229 & \cellcolor{green1!55}0.234 \\
\graymidrule
\multirow{4}{*}{\rotatebox[origin=c]{90}{\parbox{52pt}{\centering\footnotesize Dense$+$sparse\\w. 3 systems}}} & BM25 + SPLADE{\sub{fr-base}} + DPR{\sub{fr-base}} & \cellcolor{red1!30}0.228 & \cellcolor{green1!55}0.267  & \cellcolor{green1!55}0.297 & \cellcolor{green1!55}0.301$^{\dagger}$ & \cellcolor{green1!55}0.296 & \cellcolor{green1!55}0.310$^{\dagger}$ & \cellcolor{green1!55}0.263 & \cellcolor{green1!55}0.287 \\
& BM25 + SPLADE{\sub{fr-base}} + ColBERT{\sub{fr-base}} & \cellcolor{green1!55}0.260  & \cellcolor{green1!55}0.281  & \cellcolor{green1!55}0.308$^{\dagger}$ & \cellcolor{green1!55}0.308$^{\dagger}$ & \cellcolor{green1!55}0.300$^{\dagger}$ & \cellcolor{green1!55}0.314$^{\dagger}$ & \cellcolor{green1!55}0.266 & \cellcolor{green1!55}0.282 \\
& BM25 + DPR{\sub{fr-base}} + ColBERT{\sub{fr-base}} & \cellcolor{green1!55}0.238  & \cellcolor{green1!55}0.289  & \cellcolor{green1!55}0.302$^{\dagger}$ & \cellcolor{green1!55}0.308$^{\dagger}$ & \cellcolor{green1!55}0.287 & \cellcolor{green1!55}0.314$^{\dagger}$ & \cellcolor{green1!55}0.257 & \cellcolor{green1!55}0.263 \\
& SPLADE{\sub{fr-base}} + DPR{\sub{fr-base}} + ColBERT{\sub{fr-base}} & \cellcolor{red1!30}0.226 & \cellcolor{red1!30}0.232 & \cellcolor{red1!30}0.229 & \cellcolor{green1!55}0.250 & \cellcolor{red1!30}0.229 & \cellcolor{green1!55}0.249 & \cellcolor{red1!30}0.212 & \cellcolor{green1!55}0.233 \\
\graymidrule
\multirow{1}{*}{\rotatebox[origin=c]{90}{\footnotesize All}} & BM25 + SPLADE{\sub{fr-base}} + DPR{\sub{fr-base}} + ColBERT{\sub{fr-base}}& \cellcolor{green1!55}0.254 & \cellcolor{green1!55}0.275  & \cellcolor{green1!55}0.307$^{\dagger}$ & \cellcolor{green1!55}0.315$^{\dagger}$ & \cellcolor{green1!55}0.300$^{\dagger}$ & \cellcolor{green1!55}0.323$^{\dagger}$ & \cellcolor{green1!55}0.260 & \cellcolor{green1!55}0.277 \\
\bottomrule
\end{tabular}
}
\caption{Out-of-domain recall\at10 results on LLeQA dev set. We report performance of normalized score fusion using both equal and tuned weights between systems. Hybrid combinations that improve over each of their constituent systems are highlighted in \colorbox{green1!55}{green}, while those that underperform compared to one or more of their systems are marked in \colorbox{red1!30}{red}. $\dagger$ indicates competitive performance with state-of-the-art BGE-M3{\sub{dense}} (30.6\% R\at10).}
\label{tab:outdomain_fusion}
\end{table*}
\setlength{\tabcolsep}{6pt}

\paragraph{How complementary are distinct retrievers?}
We select the two systems that showed the best hybrid sparse-dense performance in \Cref{tab:outdomain_fusion}, namely BM25+ColBERT{\sub{fr-base}}, and analyze their min-max scaled scores across 18.6K query-article pairs from LLeQA, balanced between positive and negative instances. We examine four scenarios: (A) BM25 scores high (above the third quartile of its distribution, depicted in \Cref{fig:normalized_distributions}) while ColBERT{\sub{fr-base}} scores low (below the first quartile of its distribution); (B) BM25 scores low while ColBERT{\sub{fr-base}} scores high; (C) both systems score high; (D) both systems score low. Our findings, shown in \Cref{fig:complementarity}, reveal that when one system scores high while the other does not, the higher-scoring system generally provides the correct signal, effectively compensating for the other's error. Conversely, when both systems concur on the relevance assessment, whether high or low, they are predominantly correct.
%18.6K query-article pairs from LLeQA training set

\paragraph{Does fusion always help for OOD data?}
We conduct an exhaustive evaluation across all possible combinations of our learned retrievers (excluding the monoBERT{\sub{fr-base}} re-ranker due to its high inefficiency for end-to-end retrieval) and BM25, using the fusion methods presented in \Cref{sec:methodology}. For NSF, we test both conventional min-max and z-score scaling, as well as our proposed percentile normalization, with either equal or tuned weights. This results in a total of 88 different configurations, whose results are presented in \Cref{tab:outdomain_fusion}. Of these, we find that 72 (i.e., 82\%) improve performance compared to using the retrievers from the respective combinations individually. Remarkably, nine combinations outperform the extensively trained BGE-M3{\sub{dense}} model, which demonstrates the best individual performance by far on LLeQA dev set. Overall, our findings indicate that fusion \textsl{almost} always enhance performance on out-of-distribution data, regardless of the fusion technique or normalization approach used -- though tuned NSF with z-score scaling seems to deliver optimal results.

\section{In-Domain Evaluation \label{sec:indomain_eval}}
We now investigate the performance enhancement given by specialized retrievers trained on the legal domain and assess the effectiveness of fusion techniques in this in-domain context. Specifically, we explore the following question: \textsl{Assuming a limited amount of domain-specific labeled data, to what extent can specialized retrievers and their fusion enhance performance within the legal domain?} To address this question, we fine-tune our domain-general neural retrievers, initially trained on mMARCO-fr, on the 1.5K training questions from LLeQA. We denote the resulting models with the \textsc{fr-lex} subscript in the remainder of the paper.

\begin{table*}[t]
  \begin{minipage}[t]{0.475\textwidth}
    \vspace{0pt}
    \centering
    \resizebox{\columnwidth}{!}{%
    \begin{tabular}{ll|ccccc} 
    \toprule
    &\textbf{Model} & \textbf{R\at1k} & \textbf{R\at500} & \textbf{R\at100} & \textbf{R\at10} & \textbf{RP} \\
    \midrule
    \multirow{5}{*}{\rotatebox[origin=c]{90}{Dev}} & BM25 & 0.634 & 0.577 & 0.457 & 0.232 & 0.122 \\ 
    & SPLADE{\sub{fr-lex}} & 0.925 & 0.889 & 0.792 & \underline{0.535} & \underline{0.334} \\
    \graypartialmidrule{2-7}
    & DPR{\sub{fr-lex}} & \underline{0.948} & \underline{0.927} & \textbf{0.855} & \textbf{0.595} & \textbf{0.462} \\
    & ColBERT{\sub{fr-lex}} & 0.892 & 0.852 & 0.747 & 0.434 & 0.255 \\
    \graypartialmidrule{2-7}
    & monoBERT{\sub{fr-lex}} & \textbf{0.967} & \textbf{0.942} & \underline{0.805} & 0.430 & 0.219 \\
    \midrule
    \multirow{5}{*}{\rotatebox[origin=c]{90}{Test}} & BM25 & 0.742 & 0.672 & 0.537 & 0.367 & \underline{0.163} \\ 
    & SPLADE{\sub{fr-lex}} & 0.903 & 0.857 & 0.687 & 0.434 & 0.102 \\
    \graypartialmidrule{2-7}
    & DPR{\sub{fr-lex}} & \underline{0.937} & \underline{0.916} & \textbf{0.801} & \textbf{0.558} & \textbf{0.244} \\
    & ColBERT{\sub{fr-lex}} & 0.841 & 0.800 & 0.679 & 0.432 & 0.125 \\
    \graypartialmidrule{2-7}
    & monoBERT{\sub{fr-lex}} & \textbf{0.980} & \textbf{0.939} & \underline{0.746} & \underline{0.473} & 0.143 \\
    \bottomrule
    \end{tabular}
    }
    \caption{In-domain performance on LLeQA dev and test sets. We train each model five times with different seeds and report the best based on the dev set results.}
    \label{tab:indomain_lleqa}
  \end{minipage}
  \hfill
  \begin{minipage}[t]{0.5\textwidth}
    \vspace{0pt}
    \centering
    \resizebox{\columnwidth}{!}{%
    \begin{tabular}{l|cc|ccc|r} 
    \toprule
    \textbf{Model} & \multicolumn{2}{c|}{\textbf{Recall at cut-off $k$}} & \textbf{$\Delta$ Avg.} \\ 
    \midrule
    \multicolumn{1}{c}{} & \textbf{\at1000} & \multicolumn{1}{c}{\textbf{\at500}} & \\
    DPR{\sub{fr-lex}}      & \colorbox{RedOrange!15}{0.925} / \colorbox{RoyalBlue!15}{0.933} & \colorbox{RedOrange!15}{0.888} / \colorbox{RoyalBlue!15}{0.905} & +1.3\%\\
    SPLADE{\sub{fr-lex}}   & \colorbox{RedOrange!15}{0.863} / \colorbox{RoyalBlue!15}{0.878} & \colorbox{RedOrange!15}{0.817} / \colorbox{RoyalBlue!15}{0.821} & +1.0\%\\
    ColBERT{\sub{fr-lex}}  & \colorbox{RedOrange!15}{0.806} / \colorbox{RoyalBlue!15}{0.835} & \colorbox{RedOrange!15}{0.777} / \colorbox{RoyalBlue!15}{0.806} & +2.9\%\\
    monoBERT{\sub{fr-lex}} & \colorbox{RedOrange!15}{0.967} / \colorbox{RoyalBlue!15}{0.967} & \colorbox{RedOrange!15}{0.928} / \colorbox{RoyalBlue!15}{0.927} & -0.1\%\\
    \midrule
    \multicolumn{1}{c}{} & \textbf{\at50} & \multicolumn{1}{c}{\textbf{\at10}} & \\
    DPR{\sub{fr-lex}}      & \colorbox{RedOrange!15}{0.685} / \colorbox{RoyalBlue!15}{0.706} & \colorbox{RedOrange!15}{0.526} / \colorbox{RoyalBlue!15}{0.541} & +1.8\%\\
    SPLADE{\sub{fr-lex}}   & \colorbox{RedOrange!15}{0.617} / \colorbox{RoyalBlue!15}{0.596} & \colorbox{RedOrange!15}{0.402} / \colorbox{RoyalBlue!15}{0.403} & -1.0\%\\
    ColBERT{\sub{fr-lex}}  & \colorbox{RedOrange!15}{0.593} / \colorbox{RoyalBlue!15}{0.599} & \colorbox{RedOrange!15}{0.388} / \colorbox{RoyalBlue!15}{0.416} & +1.7\%\\
    monoBERT{\sub{fr-lex}} & \colorbox{RedOrange!15}{0.632} / \colorbox{RoyalBlue!15}{0.629} & \colorbox{RedOrange!15}{0.353} / \colorbox{RoyalBlue!15}{0.335} & -1.2\%\\
    \bottomrule
    \end{tabular}
    }
    \caption{In-domain recall\at$k$ performances on LLeQA test set \colorbox{RedOrange!15}{\textsl{without}} / \colorbox{RoyalBlue!15}{\textsl{with}} pre-finetuning on mMARCO-fr. We report the means across 5 runs with different seeds.}
    \label{tab:effect_prefinetuning}
  \end{minipage}
\end{table*}

\begin{figure*}[t]
\centering
\begin{subfigure}[t]{.33\textwidth}
  \centering
  \includegraphics[width=\linewidth]{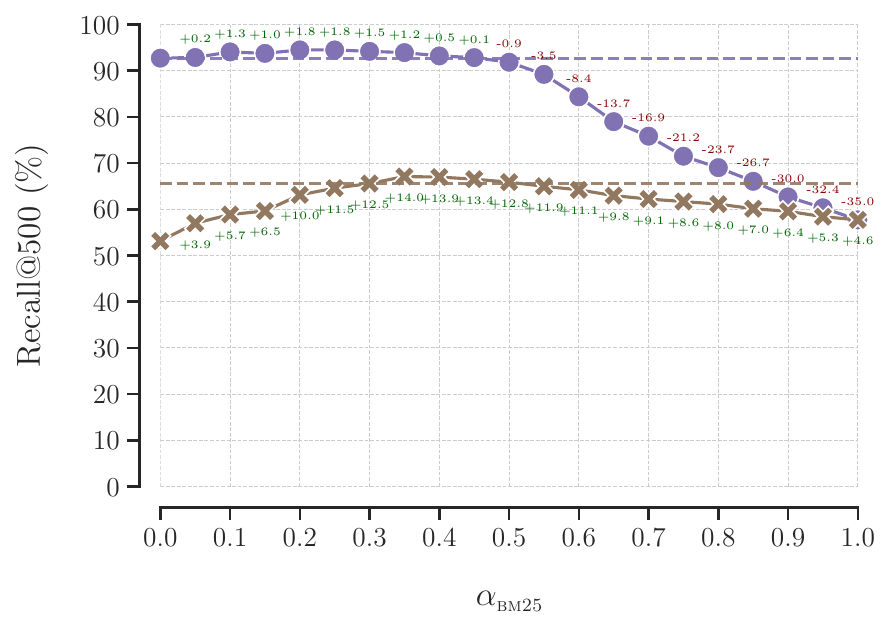}
\end{subfigure}
\begin{subfigure}[t]{.33\textwidth}
  \centering
  \includegraphics[width=\linewidth]{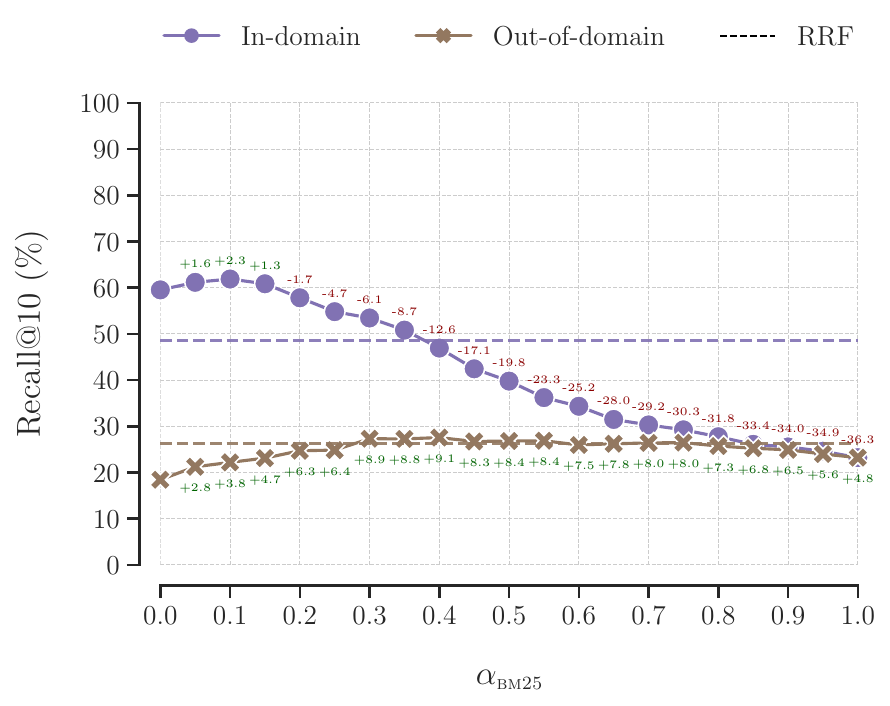}
\end{subfigure}
\begin{subfigure}[t]{.329\textwidth}
  \centering
  \includegraphics[width=\linewidth]{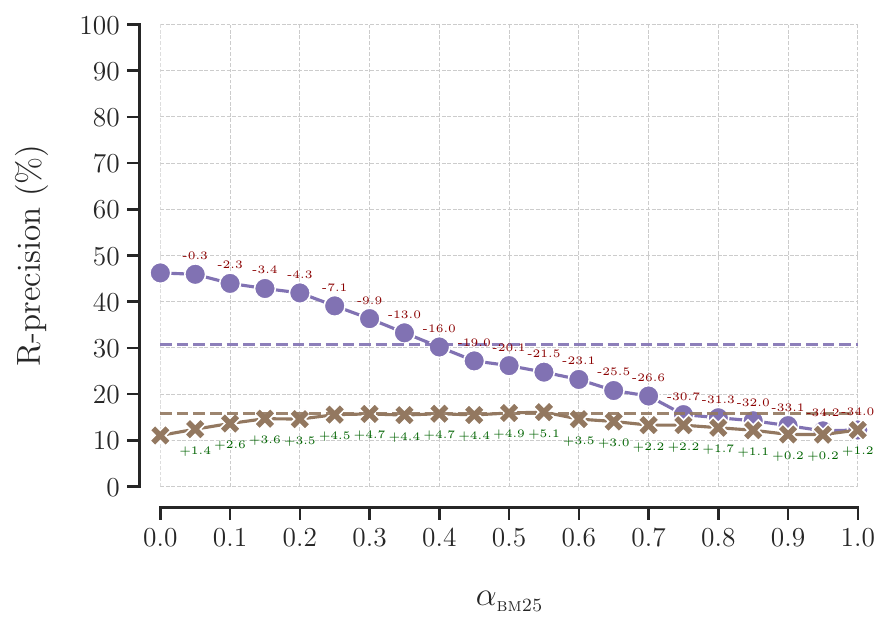}
\end{subfigure}
\caption{Effect of weight tuning in normalized score fusion between BM25 and DPR{\sub{fr-\{\textcolor[HTML]{8172b3}{\textbf{lex}},\textcolor[HTML]{937860}{\textbf{base}}\}}} on LLeQA dev set.}
\label{fig:effect_weight_tuning}
\end{figure*}

\paragraph{Main results.}
\Cref{tab:indomain_lleqa} presents the in-domain performance of our specialized retrieval models. In line with previous findings \citep{karpukhin2020dense, khattab2020colbert, formal2021splade, nogueira2019multi}, we note substantial improvements across all models compared to the zero-shot setting, with each now significantly outperforming the robust BM25 baseline. Interestingly, our single-vector dense retriever, DPR{\sub{fr-lex}}, surpasses all the other approaches, including the more computationally demanding monoBERT{\sub{fr-lex}} cross-encoder on smaller recall cutoffs. These results underscore the effectiveness of neural methods when trained in-domain, even with relatively limited sample sizes.

\paragraph{Is task-adaptive pre-finetuning beneficial?}
Here, we study the hypothesis that performing an intermediary finetuning step on a task-related dataset before finetuning on the target dataset can help enhance downstream performance \citep{dai2019deeper,li2020parade}, especially when training samples in the target domain are scarce \citep{zhang2020little}. We therefore compare two learning strategies: the first directly finetunes the pretrained CamemBERT backbone on the specialized LLeQA dataset, while the second (which we adopted as our default approach) incorporates a pre-finetuning step on the domain-general mMARCO-fr dataset. We find this intermediary phase to consistently improve in-domain performance at higher recall cutoffs across all bi-encoder models, as shown in \Cref{tab:effect_prefinetuning}. However, this benefit appears limited to dense representation models at lower recall cutoffs, with SPLADE{\sub{fr-lex}} experiencing diminished performance. As for the monoBERT{\sub{fr-lex}} cross-encoder, pre-finetuning does not yield improvements.
%initially pre-finetuned on the domain-general mMARCO-fr dataset and subsequently adapted to the legal domain via training on LLeQA.

\setlength{\tabcolsep}{7pt}
\begin{table*}[t]
\centering
\resizebox{\textwidth}{!}{%
\begin{tabular}{l|c|c|ll|ll|ll} 
\toprule
\multirow{2}{*}{\textbf{Method}} & \multirow{2}{*}{\textbf{BCF}} & \multirow{2}{*}{\textbf{RRF}} & \multicolumn{2}{c|}{\textbf{NSF{\sub{min-max}}}} & \multicolumn{2}{c|}{\textbf{NSF{\sub{z-score}}}} & \multicolumn{2}{c}{\textbf{NSF{\sub{percentile}}}} \\
&&& Equal & Tuned & Equal & Tuned & Equal & Tuned \\
\midrule\midrule
BM25                    & 0.232 & 0.232 & 0.232 & 0.232 & 0.232 & 0.232 & 0.232 & 0.232 \\
DPR{\sub{fr-lex}}      & 0.595 & 0.595 & 0.595 & 0.595 & 0.595 & 0.595 & 0.595 & 0.595 \\
SPLADE{\sub{fr-lex}}   & 0.535 & 0.535 & 0.535 & 0.535 & 0.535 & 0.535 & 0.535 & 0.535 \\
ColBERT{\sub{fr-lex}}  & 0.434 & 0.434 & 0.434 & 0.434 & 0.434 & 0.434 & 0.434 & 0.434 \\
\midrule
BM25 + SPLADE{\sub{fr-lex}} & \cellcolor{red1!30}0.385 & \cellcolor{red1!30}0.457 & \cellcolor{red1!30}0.417 & \cellcolor{green1!55}0.570 & \cellcolor{red1!30}0.350 & \cellcolor{green1!55}0.561 & \cellcolor{red1!30}0.369 & \cellcolor{red1!30}0.450 \\
\graymidrule
DPR{\sub{fr-lex}} + ColBERT{\sub{fr-lex}} & \cellcolor{red1!30}0.546 & \cellcolor{red1!30}0.541 & \cellcolor{red1!30}0.577 & \cellcolor{green1!55}0.609$^{\dagger}$ & \cellcolor{red1!30}0.592 & \cellcolor{green1!55}0.608$^{\dagger}$ & \cellcolor{red1!30}0.464 & \cellcolor{red1!30}0.555\\
\graymidrule
BM25 + DPR{\sub{fr-lex}} & \cellcolor{red1!30}0.391 & \cellcolor{red1!30}0.485 & \cellcolor{red1!30}0.398 & \cellcolor{green1!55}0.619$^{\dagger}$ & \cellcolor{red1!30}0.326 & \cellcolor{green1!55}0.618$^{\dagger}$ & \cellcolor{red1!30}0.351 & \cellcolor{red1!30}0.452 \\
BM25 + ColBERT{\sub{fr-lex}} & \cellcolor{red1!30}0.363 & \cellcolor{red1!30}0.412 & \cellcolor{red1!30}0.360 & \cellcolor{green1!55}0.470 & \cellcolor{red1!30}0.288 & \cellcolor{green1!55}0.473 & \cellcolor{red1!30}0.383 & \cellcolor{red1!30}0.437 \\
SPLADE{\sub{fr-lex}} + DPR{\sub{fr-lex}} & \cellcolor{red1!30}0.573 & \cellcolor{red1!30}0.586 & \cellcolor{red1!30}0.582 & \cellcolor{green1!55}0.613$^{\dagger}$ & \cellcolor{red1!30}0.586 & \cellcolor{green1!55}0.612$^{\dagger}$ & \cellcolor{red1!30}0.587 & \cellcolor{green1!55}0.604 \\
SPLADE{\sub{fr-lex}} + ColBERT{\sub{fr-lex}} & \cellcolor{red1!30}0.514 & \cellcolor{red1!30}0.509 & \cellcolor{green1!55}0.537 & \cellcolor{green1!55}0.557 & \cellcolor{green1!55}0.543 & \cellcolor{green1!55}0.553 & \cellcolor{red1!30}0.464 & \cellcolor{red1!30}0.519 \\
\graymidrule
BM25 + SPLADE{\sub{fr-lex}} + DPR{\sub{fr-lex}} & \cellcolor{red1!30}0.431 & \cellcolor{green1!55}0.606$^{\dagger}$ & \cellcolor{red1!30}0.533 & \cellcolor{green1!55}0.629$^{\dagger}$ & \cellcolor{red1!30}0.447 & \cellcolor{green1!55}0.625$^{\dagger}$ & \cellcolor{red1!30}0.395 & \cellcolor{red1!30}0.472 \\
BM25 + SPLADE{\sub{fr-lex}} + ColBERT{\sub{fr-lex}} & \cellcolor{red1!30}0.427 & \cellcolor{red1!30}0.535 & \cellcolor{red1!30}0.505 & \cellcolor{green1!55}0.575 & \cellcolor{red1!30}0.402 & \cellcolor{green1!55}0.578 & \cellcolor{red1!30}0.412 & \cellcolor{red1!30}0.475 \\
BM25 + DPR{\sub{fr-lex}} + ColBERT{\sub{fr-lex}} & \cellcolor{red1!30}0.429 & \cellcolor{red1!30}0.564 & \cellcolor{red1!30}0.481 & \cellcolor{green1!55}0.624$^{\dagger}$ & \cellcolor{red1!30}0.372 & \cellcolor{green1!55}0.623$^{\dagger}$ & \cellcolor{red1!30}0.402 & \cellcolor{red1!30}0.468 \\
SPLADE{\sub{fr-lex}} + DPR{\sub{fr-lex}} + ColBERT{\sub{fr-lex}} & \cellcolor{red1!30}0.548 & \cellcolor{red1!30}0.579 & \cellcolor{red1!30}0.579 & \cellcolor{green1!55}0.617$^{\dagger}$ & \cellcolor{red1!30}0.587 & \cellcolor{green1!55}0.620$^{\dagger}$ & \cellcolor{red1!30}0.480 & \cellcolor{red1!30}0.560 \\
\graymidrule
BM25 + SPLADE{\sub{fr-lex}} + DPR{\sub{fr-lex}} + ColBERT{\sub{fr-lex}} & \cellcolor{red1!30}0.457 & \cellcolor{green1!55}0.603$^{\dagger}$ & \cellcolor{red1!30}0.561 & \cellcolor{green1!55}0.628$^{\dagger}$ & \cellcolor{red1!30}0.485 & \cellcolor{green1!55}0.627$^{\dagger}$ & \cellcolor{red1!30}0.418 & \cellcolor{red1!30}0.477 \\
\bottomrule
\end{tabular}
}
\caption{In-domain recall\at10 results on LLeQA dev set. The \colorbox{red1!30}{red} region highlights hybrid combinations that perform worse than one or more of their systems, while the \colorbox{green1!55}{green} region emphasizes combinations that outperform each of their constituent systems. $\dagger$ indicates improved performance over DPR{\sub{fr-lex}} alone.}
\label{tab:indomain_fusion}
\end{table*}
\setlength{\tabcolsep}{6pt}

\paragraph{Does fusion still help with specialized retrievers?}
\Cref{tab:indomain_fusion} highlights the in-domain performance of hybrid combinations previously assessed in a zero-shot setting. We now observe a very distinct pattern: around 70\% of these combinations lead to deteriorated performance compared to using one of their constituent systems only. Among the 27 (out of 88) configurations that do show improvement, 23 leverage NSF with weights tuned in-domain, while only four combinations (i.e., 5\% in total) achieve superior performance without prior tuning. Furthermore, the performance gap between individual systems and their hybrid combinations is considerably narrower within this in-domain context. While a two-system hybrid fusion can yield up to a 7.1\% R\at10 improvement over the best single system in zero-shot scenarios, this enhancement does not exceed 1.4\% once the models are trained in-domain. \Cref{app:indomain_complentarity} further discusses that degradation.

\paragraph{How does $\alpha$ in paired NSF affect performance?}
Finally, we evaluate the impact of weight tuning on the in-domain performance of NSF in a paired configuration, where one system is assigned a weight $\alpha$ and the other $1\!-\!\alpha$. We select the best performing two-system combination from \Cref{tab:indomain_fusion}, i.e., BM25+DPR{\sub{fr-lex}}. For comparison, we also report performance of this combination in a zero-shot context and that of RRF in both scenarios, as depicted in \Cref{fig:effect_weight_tuning}. We find that integrating BM25 offers minimal benefits once DPR{\sub{fr}} is domain-tuned, with equal weighting between both systems consistently leading to worse performance. This finding contrasts starkly with the out-of-distribution setting, where combining both systems consistently improves performance compared to using one of them alone, regardless of the $\alpha$ weight assigned.

% -------------------------------------------------------------------------
%                             RELATED WORK
% -------------------------------------------------------------------------
\section{Related Work \label{sec:related_work}}

\paragraph{Statute law retrieval.}
Returning the relevant legislation to a short legal question is notably challenging due to the linguistic disparity between the specialized jargon of legal statutes \citep{charrow1978legal} and the plain language typically used by laypeople. Research on statute retrieval has traditionally focused on text-level similarity between queries and candidate documents, with earlier methods employing lexical approaches such as TF-IDF \citep{kim2017two, dang2019approach} or BM25 \citep{wehnert2019threshold, gain2019iitp}. With advancements in representation learning techniques \citep{vaswani2017attention, devlin2019bert}, attention has shifted towards dense retrieval to enhance semantic matching capabilities. For instance, \citet{louis2022statutory} demonstrate that supervised single-vector dense bi-encoders significantly outperform TF-IDF weighting schemes. \citet{su2024stard} explore various dense bi-encoder models trained on different domains and reached similar conclusions. \citet{santosh2024cusines} further push performance of dense bi-encoders by introducing a dynamic negative sampling strategy tailored to law. In parallel, some studies have begun incorporating legal knowledge into the retrieval process. For example, \citet{louis2023finding} propose a graph-augmented dense retriever that uses the topological structure of legislation to enrich article content information. Meanwhile, \citet{qin2024explicitely} develop a generative model that learns to represent legal documents as hierarchical semantic IDs before associating queries with their relevant document IDs. Despite this progress, no studies have explored the potential of combining diverse retrieval approaches in the legal domain, especially in zero-shot settings using domain-general models, which may individually struggle due to the specialized nature of law.
%Legal texts are indeed characterized by their extensive use of formal vocabulary, Latin phrases, extended sentences, and terms with variable meanings, which complicates the matching process.

\paragraph{French language representation.}
Existing research in NLP predominantly focuses on English-centric directions \citep{arr2024linguistic}. In French, efforts have been made in developing monolingual pretrained language models in various configurations: encoder-only \citep{martin2020camembert, le2020flaubert, antoun2023data}, seq2seq \citep{eddine2021barthez}, and decoder-only \citep{louis2020belgpt, simoulin2021modele, muller2022cedille, launay2022pagnol}. Despite these advancements, specialized models for French remain scarce, largely due to the limited availability of high-quality labeled data. This scarcity is particularly pronounced in the field of retrieval, with few exceptions \citep{arbaretier2024solon}. As a result, practitioners typically rely on larger multilingual models \citep{wang2024me5, chen2024bge} that distribute tokens and parameters across various languages, often leading to sub-optimal downstream performance due to the curse of multilinguality \citep{conneau2020unsupervised}.

%-------------------------------------------------------------------------
%                             CONCLUSION
%-------------------------------------------------------------------------
\section{Conclusion \label{sec:conclusion}}
Our work explores the potential of combining distinct retrieval methods in a non-English specialized domain, specifically French statute laws. Our findings reveal that supervised domain-general monolingual models, trained with limited resources, can rival leading multilingual retrieval models, though are more vulnerable to out-of-distribution data. However, combining these monolingual models almost consistently enhances their zero-shot performance, regardless of the fusion technique employed, with certain combinations achieving state-of-the-art results in the legal domain. We show the complementary nature of these models and find they can effectively compensate each other's mistakes, hence the performance boost. Furthermore, we confirm that in-domain training significantly enhances the effectiveness of neural retrieval models, while pre-finetuning can help with dense bi-encoders. Finally, our results indicate that fusion generally does not benefit specialized retrievers and only improves performance when scores are fused with carefully tuned weights, as equal weighting consistently leads to reduced performance. Overall, these insights suggest that for specialized domains, finetuning a single bi-encoder generally yields optimal results when (even limited) high-quality domain-specific data is available, whereas fusion should be preferred when such data is not accessible and domain-general retrievers are used.

% - Domain-general monolingual retrieval models show competitive in-domain performance with state-of-the-art multilingual models on domain-general data while trained with a fraction of the costs.
% - However they underperform compared to these extensively trained multilingual models on out-of-distribution data, though the naturally zero-shot BM25 baseline achieves the best results.
% - However fusing our learned retrievers  compared to using the models alone, and some combinations with BM25 even lead to sota performance.
% - We show that sparse and dense models are indeed complementary as to how they moel relevance and that that they can effectively compensate for th other's error when fused
% - We show that the same nural models trained in-domain significantly improve performance compared to the zero-shot setting.
% - We demonstrate the pre-finetuning can help downstream performance for some models.
% - We show that in contrast with the zero-shot setting, fusion rarely helps when the models have been trained in-domain, and when it does it is after tuning weights for a score fusion.
% - We visualize this behavior for different values of alpha and compare it to the zero-shot setting.

%-------------------------------------------------------------------------
%                             LIMITATIONS
%-------------------------------------------------------------------------
\section*{Limitations \label{sec:limitations}}
We identify three core limitations in our research.

Firstly, our analysis specifically targets two underexplored areas -- the legal domain and the French language -- and is therefore confined to the only dataset available in this niche (LLeQA; \citealp{louis2024lleqa}). This raises questions about the generalizability of our findings across broader French legal resources, such laws from different French-speaking jurisdictions (e.g., France, Switzerland, or Canada) or across legal topics beyond those covered in LLeQA.

Secondly, our study focuses solely on end-to-end retrievers -- i.e., systems that identify and fetch all potentially relevant items from an entire knowledge corpus -- as opposed to ranking methods that take the output of retrievers and sort it. Specifically, we deliberately omit the monoBERT{\sub{fr}} ranker due to its prohibitive inference costs for end-to-end retrieval -- a brute-force search across all 28K articles in LLeQA requires about two minutes per query on GPU, a latency 9500$\times$ higher than that of single-vector retrieval, making it impractical for real-world retrieval. We let the exploration of fusion with re-rankers for future work.
%Fusion with cross-encoders: https://arxiv.org/pdf/2301.09728, https://arxiv.org/pdf/2201.07667, https://ceur-ws.org/Vol-2950/paper-02.pdf

Lastly, although beyond the scope of our work, it remains an open question whether the present findings are applicable to other non-English languages within different highly specialized domains.

%-------------------------------------------------------------------------
%                             ETHICS STATEMENT
%-------------------------------------------------------------------------
\section*{Ethical Considerations \label{sec:ethics}}
The scope of this work is to drive research forward in legal information retrieval by uncovering novel insights on fusion strategies. We believe this is an important application field where more research could improve legal aid services and access to justice for all. We do not foresee major situations where our methodology and findings would lead to harm \citep{tsarapatsanis2021ethical}. Nevertheless, we emphasize that the premature deployment of prominent retrieval models not tailored for the legal domain poses a tangible risk to laypersons, who may uncritically rely on the provided information when faced with a legal issue and inadvertently worsen their personal situations.

%-------------------------------------------------------------------------
%                             ACKNOWLEDGMENTS
%-------------------------------------------------------------------------
% \section*{Acknowledgments}
% This research is partially supported by the Sector Plan Digital Legal Studies of the Dutch Ministry of Education, Culture, and Science. In addition, this research was made possible, in part, using the Data Science Research Infrastructure (DSRI) hosted at Maastricht University.

%-------------------------------------------------------------------------
%                             BIBLIOGRAPHY
%-------------------------------------------------------------------------
% Entries for the entire Anthology, followed by custom entries
\bibliographystyle{packages/acl_natbib}
\bibliography{refs.bib}

%-------------------------------------------------------------------------
%                             APPENDIX
%-------------------------------------------------------------------------
\appendix
\section{Methodology Details\label{app:methodology_details}}
Formally speaking, a statutory article retrieval system takes as input a question $q$ along with a corpus of law articles $\mathcal{C}$, and returns a ranked list $\mathcal{R}_q \subset \mathcal{C}$ of the supposedly relevant articles, sorted by decreasing order of relevance.

\begin{figure*}[t]
    \centering
    \includegraphics[width=1.0\linewidth]{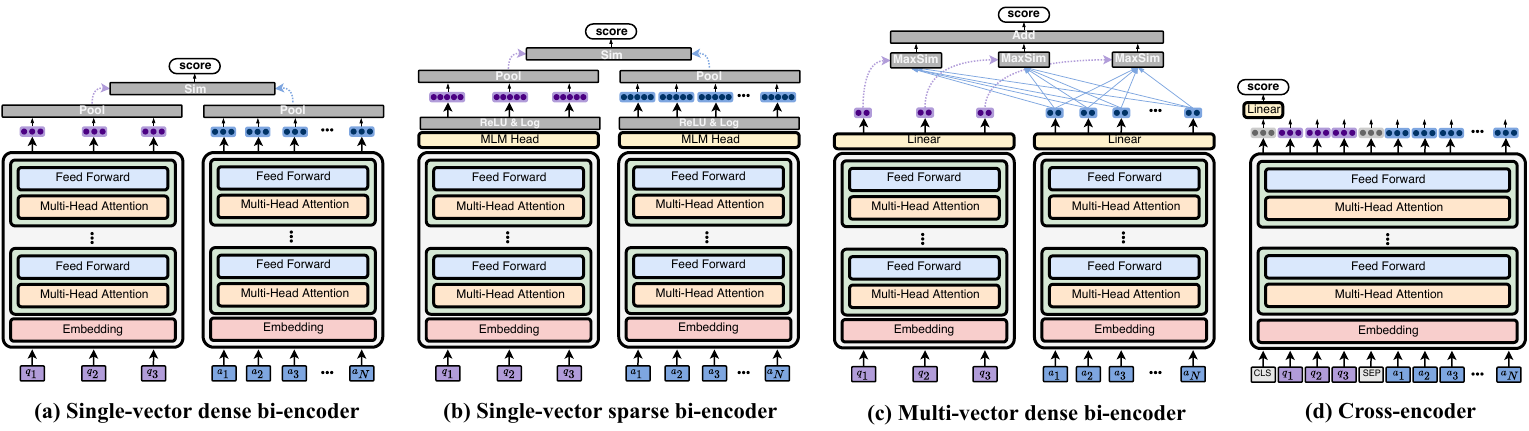}
    \caption{High-level illustration of the four prominent neural retrieval architectures explored in this study.}
    \label{fig:models}
\end{figure*}

\subsection{Retrieval Models\label{app:retrieval_models}}

\noindent\textbf{BM25} \citep{robertson1994okapi}\quad is an unsupervised probabilistic weighting scheme that estimates relevance based on term-matching between high-dimensional sparse vectors using statistical properties such as term frequencies, document frequencies, and document lengths. Specifically, it calculates a relevance score $s(q,a): \mathcal{V}^{|q|} \times \mathcal{V}^{|a|} \rightarrow \mathbb{R}_+$ between query $q$ and article $a$ as a sum of contributions of each query term $t$ from vocabulary $\mathcal{V}$ appearing in the article, i.e.,
\begin{equation}
\label{eq:score_bm25}
\begin{split}
s_{\textsc{bm25}}\!\left(q, a\right) = \sum_{t \in q}
    &\log\!\left( \frac{|\mathcal{C}|-\operatorname{df}(t)+0.5}{\operatorname{df}(t)+0.5}\right)\\
    &{\cdot}\frac{\operatorname{tf}(t, a){\cdot} (k_1{+}1)}{\operatorname{tf}(t, a){+}k_1{\cdot}\left(1{-}b{+}b\cdot \frac{|a|}{avgal}\right)},
\end{split}
\end{equation}
where the term frequency $\operatorname{tf}(t, a): \mathcal{V}^1 \times \mathcal{V}^{|a|} \rightarrow \mathbb{Z}_+$ is the number of occurrences of term $t$ in article $a$, the document frequency $\operatorname{df}(t): \mathcal{V}^1 \rightarrow \mathbb{Z}_+$ is the number of articles within the corpus $\mathcal{C}$ that contain term $t$, $k_1 \in \mathbb{R}_+$ and $b \in [0,1]$ are constant parameters, and $avgal$ is the average article length.\\
\noindent BM25 remains widely used due to its balance between simplicity and robustness, often competing with modern retrieval methods \citep{thakur2021beir} while being extremely efficient and requiring no training. However, its reliance on exact-term matching restricts its ability to understand semantics, capture contextual relationships, and handle synonyms or rare terms.

%\paragraph{DPR{\sub{fr}}} is a supervised \textsl{single-vector dense} bi-encoder model \citep{lee2019latent,chang2020pretraining,karpukhin2020dense} that estimates relevance via cosine similarity between independent query and article latent representations. 
\paragraph{DPR{\sub{fr-\{base,lex\}}}} are based on the widely-adopted siamese bi-encoder architecture \citep{gillick2018end}, which consists of a learnable text embedding function $E(i; \boldsymbol{\Omega}): \mathcal{V}^{n} \mapsto \mathbb{R}^{n \times d}$ that maps an input text sequence $i$ of $n$ terms from vocabulary $\mathcal{V}$ to $d$-dimensional real-valued term vectors, i.e.,
\begin{equation}
\label{eq:func_encoder}
     E\!\left(i;\boldsymbol{\Omega} \right) = \mathbf{H}_i = \left[\mathbf{h}_{i,\textsc{cls}}, \mathbf{h}_{i,1}, \cdots, \mathbf{h}_{i,n}\right],
\end{equation}
and calculates a relevance score between query $q$ and article $a$ by operating on their independently computed bags of contextualized term embeddings $\mathbf{H}_i \in \mathbb{R}^{n \times d}$. Our single-vector dense representation models obtain this score by performing
\begin{equation}
\label{eq:score_single}
    s_{\textsc{single}}\!\left(q, a\right) = \mathbf{h}^*_q \cdot \mathbf{h}^*_a,
\end{equation}
where $\mathbf{h}^*_i \in \mathbb{R}^d$ is the global representation of sequence $i$, derived by mean pooling across the sequence term embeddings, i.e.,
\begin{equation}
\label{eq:avg_pooling}
    \mathbf{h}^*_i = \operatorname{AvgP}\!\left(\mathbf{H}_i\right) = \frac{1}{|i|} \mathbf{H}_{i}^\mathsf{T} \mathbf{1}_{|i|}.
\end{equation}
The models are trained via optimization of the contrastive NT-Xent loss \citep{chen2020simclr,gao2021simcse}, which aims to learn a high-quality embedding function so that relevant query-article pairs achieve higher similarity than irrelevant ones. Let $\mathcal{B}=\{(q_i, a^{+}_i, a_{\textsc{h},i}^{-})\}_{i=1}^{N}$ be a batch of $N$ training instances, each comprising a query $q_i$ associated with a positive article $a^{+}_i$ and a hard negative article $a_{\textsc{h},i}^{-}$. By considering the articles paired with all other queries within the same batch, we can enrich each training triple with an additional set of 2($N-$1) in-batch negatives $\mathcal{A}_{\textsc{ib},i}^{-} = \{a^{+}_j,a_{\textsc{h},j}^{-}\}_{j \neq i}^{N}$. Given these augmented training samples, we contrastively optimize the negative log-likelihood of each positive article such that
\begin{equation}
\label{eq:loss_ntxent}
    \mathcal{L}_{\textsc{nt-xent}} = -\log \frac{e^{s(q_i,a^{+}_i)/\tau}}
{\sum_{a \in \{a^{+}_i,a_{\textsc{h},i}^{-}\} \cup \mathcal{A}_{\textsc{ib},i}^{-}} e^{\operatorname{s}(q_i, a)/\tau}},
\end{equation}
where $\tau \in \mathbb{R}_+$ is a temperature hyper-parameter that controls the concentration level of the distribution \citep{hinton2015distilling}. We enforce 
$\|\mathbf{h}^*_i\| = 1$ via a $\ell_2$-normalization layer such that \Cref{eq:score_single} computes the cosine similarity.\\
\noindent Single-vector dense models proved to effectively model language nuances and contextual information \citep{karpukhin2020dense}. Furthermore, the independent encoding enables offline pre-computation of article embeddings, resulting in low latency query-time retrieval. However, its effectiveness can be limited by the quality and diversity of its training data, potentially leading to sub-optimal performance with out-of-distribution content \citep{thakur2021beir}.

%\paragraph{SPLADE{\sub{fr}}} is a supervised \textsl{single-vector sparse} bi-encoder model \citep{formal2021spladev2, formal2021splade} that uses the MLM head of a pretrained autoencoding language model to generate term-level, expansion-aware high-dimensional representations.
\paragraph{SPLADE{\sub{fr-\{base,lex\}}}} follow SPLADE-max \citep{formal2021spladev2}, which uses the same single-vector scoring mechanism as its dense representation counterpart, outlined in \Cref{eq:score_single}, but operates on different global sequence representations derived as follows:
\begin{equation}
\mathbf{h}^*_i = \operatorname{MaxP}\!\left(\operatorname{sat}\!\left(\operatorname{transf}(\mathbf{H}_i)\mathbf{W}_{\textsc{mlm}}^\mathsf{T}\!+\!\mathbf{b}_{\textsc{mlm}})\right)\right),
\end{equation}
where $\operatorname{transf}(\cdot;\boldsymbol{\gamma}): \mathbb{R}^{n \times d} \rightarrow \mathbb{R}^{n \times d}$ first transforms the contextualized term embeddings using
\begin{equation}
\operatorname{transf}(\cdot;\boldsymbol{\gamma}) = \operatorname{LayerNorm}(\operatorname{GELU}(\operatorname{Linear}(\cdot))),
\end{equation}
preparing them for subsequent projection onto the vocabulary space via the pretrained MLM classification head $\mathbf{W}_{\textsc{mlm}} \in \mathbb{R}^{|\mathcal{V}| \times d}$, with bias $\mathbf{b}_{\textsc{mlm}} \in \mathbb{R}^{|\mathcal{V}|}$. The function $\operatorname{sat}(\cdot): \mathbb{R}^{n \times |\mathcal{V}|} \rightarrow \mathbb{R}^{n \times |\mathcal{V}|}$ then applies ReLU to ensure positive token activations, before performing log-saturation to maintain sparsity and prevent some tokens from dominating:
\begin{equation}
\operatorname{sat}(\cdot) = \log\left(1\!+\!\operatorname{ReLU}(\cdot)\right).
\end{equation}
Finally, a max pooling operation $\operatorname{MaxP}(\cdot): \mathbb{R}^{n \times |\mathcal{V}|} \rightarrow \mathbb{R}^{|\mathcal{V}|}$ is applied to distill the global sequence representation. The model is trained by jointly optimizing the contrastive NT-Xent objective, presented in \Cref{eq:loss_ntxent}, and the FLOPS regularization loss \citep{paria2020flops}, which aims to impose sparsity on the produced embeddings while encouraging an even distribution of the non-zero elements across all the dimensions to ensure maximal speedup. This is achieved by minimizing a smooth relaxation of the average number of floating-point operations necessary to compute the dot product between two embeddings (as outlined in \Cref{eq:score_single}), defined as follows:
\begin{equation}
\label{eq:loss_flops}
    \ell_{\textsc{flops}} = \sum_{j=1}^{|\mathcal{V}|} \bar{p}_j^2 
    = \sum_{j=1}^{|\mathcal{V}|}\left(\frac{1}{|\mathcal{B}|} \sum_{i=1}^{|\mathcal{B}|} \mathbf{h}^*_{ij}\right)^2
\end{equation}
where $\bar{p}_j \approx |\mathcal{B}|^{-1} \sum_{i=1}^{|\mathcal{B}|} \mathbbm{1}[\mathbf{h}^*_{ij}\!\neq\!0]$ is the empirical estimation of the activation probability for token $t_j \in \mathcal{V}$ over a batch $\mathcal{B}$. The overall loss is given by
\begin{equation}
\label{eq:loss_splade}
    \mathcal{L}_{\textsc{splade}} = \mathcal{L}_{\textsc{nt-xent}} + \lambda_q \ell_{\textsc{flops}}^{q} + \lambda_a \ell_{\textsc{flops}}^{a},
\end{equation}
where $\lambda_i$ controls the strength of the regularization, with higher values typically encouraging the model to learn sparser representations, therefore enhancing efficiency yet often at the expense of performance. By applying separate regularization weights for queries and articles, greater emphasis can be placed on sparsity for queries, which is critical for fast inference with inverted indexes.\\
\noindent  As its representations are grounded in the encoder’s vocabulary, SPLADE enhances interpretability and facilitates explanations of observed rankings. It also exhibits strong generalization capabilities on out-of-distribution data and the sparsity of its vectors enables the use of inverted indexes for fast inference. Nevertheless, learning sparse representations in high-dimensional spaces poses specific challenges: factors such as the tokenization type or the initial distribution of MLM weights can lead to model divergence \citep{formal2023towards}.

%\paragraph{ColBERT{\sub{fr}}} is a supervised \textsl{multi-vector dense} bi-encoder model \citep{khattab2020colbert, santhanam2022colbertv2} that calculates cosine similarity across all query-article term embedding pairs. It applies max-pooling to the resulting similarity scores for each query term, before summing the maximum values to estimate overall relevance.
\paragraph{ColBERT{\sub{fr-\{base,lex\}}}} use the fine-granular late interaction scoring mechanism of ColBERT \citep{khattab2020colbert}, which calculates the similarity across all pairs of query and article token embeddings, applies max-pooling across the resulting scores for each query term, and then sum the maximum values across query terms to derive the overall relevance estimate, i.e.,
\begin{equation}
\label{eq:score_colbert}
s_{\textsc{multi}}\!\left(q, a\right) = \sum_{i=1}^{|q|} \max_{j=1}^{|a|}\mathbf{h}_{q,i} \cdot \mathbf{h}_{a,j}.
\end{equation}
We train the model by jointly optimizing two contrastive objectives, namely the pairwise softmax cross-entropy loss used in ColBERTv1, defined as
\begin{equation}
\label{eq:loss_pairwise_softmax_ce}
    \mathcal{L}_{\textsc{pairsm-ce}} = -\log \frac{e^{s(q_i,a^{+}_i)}}
{e^{a(q_i,a^{+}_i)} + e^{s(q_i,a_{\textsc{h},i}^{-})}},
\end{equation}
and the NT-Xent loss, added as an enhancement for optimizing ColBERTv2 \citep{santhanam2022colbertv2}.\\
\noindent ColBERT's fine-grained late interaction between term embeddings demonstrates greater effectiveness and robustness to out-of-distribution data compared to single-vector dense bi-encoders \citep{thakur2021beir}, while enabling result interpretability. However, its computational complexity requires sophisticated engineering schemes and low-level optimizations for efficient large-scale deployment \citep{santhanam2022plaid}.

%\paragraph{monoBERT{\sub{fr}}} is a supervised \textsl{encoder-only cross-attention} model \citep{nogueira2019passage, han2020learning, gao2021rethink} that performs all-to-all interactions across terms from concatenated query-article pairs, before deriving a relevance probability through binary classification on a special token representation.
\paragraph{monoBERT{\sub{fr-\{base,lex\}}}} exploit the encoder-only cross-attention model structure \citep{nogueira2019passage}, which uses a text embedding model similar to the one defined in \Cref{eq:func_encoder} to perform all-to-all interactions across terms from concatenated query-article pairs, before deriving a relevance score through binary classification on the pair representation, i.e.,
\begin{equation}
\begin{split}
s_{\textsc{mono}}\!\left(q, a\right) 
&= \sigma\!\left(\operatorname{transf}\!\left(\mathbf{h}^*_{\left[q;a\right]}\right)\mathbf{W}_{\text{out}}^\mathsf{T}\!+\!\mathbf{b}_{\text{out}}\right),
\end{split}
\end{equation}
where $\mathbf{h}^*_{\left[q;a\right]} \in \mathbb{R}^{d}$ is obtained through a first token pooling operation $\operatorname{FirstP}(\cdot): \mathbb{R}^{n \times d} \rightarrow \mathbb{R}^{d}$, which extracts the special \textsc{cls} token representation of the concatenated sequence:
\begin{equation}
\mathbf{h}^*_{\left[q;a\right]} = \operatorname{FirstP}\!\left(\mathbf{H}_{\left[q;a\right]}\right) = \mathbf{h}_{\left[q;a\right],\textsc{cls}}.
\end{equation}
The \textsc{cls} token embedding is then transformed with $\operatorname{transf}(\cdot;\boldsymbol{\theta}): \mathbb{R}^{d} \rightarrow \mathbb{R}^{d}$ such that
\begin{equation}
\operatorname{transf}(\cdot;\boldsymbol{\theta}) = \tanh(\operatorname{Linear}(\cdot)),
\end{equation}
before being projected to a real-valued score via a linear layer $\mathbf{W}_{\text{out}} \in \mathbb{R}^{1 \times d}$ with bias $\mathbf{b}_{\text{out}} \in \mathbb{R}^{d}$. Finally, the sigmoid function $\sigma$ bounds the resulting score to the interval $[0,1]$. The model is optimized via the binary cross-entropy training objective
\begin{equation}
\label{eq:loss_binary_ce}
\begin{split}
    \mathcal{L}_{\textsc{bce}} = 
    &- y_i \cdot \log \left(s(q_i,a_i)\right) \\
    &- (1-y_i) \cdot \log \left(1-s(q_i,a_i)\right),
\end{split}
\end{equation}
where $y_i$ is the ground-truth relevance label for query-article pair $(q_i,a_i)$.\\
\noindent The rich interaction mechanism of such a model allows to capture complex relationships and often achieve state-of-the-art performance in retrieval tasks \citep{hofstatter2020improving}. However, its high computational complexity makes it impractical for large-scale or real-time retrieval scenarios, limiting its use to re-ranking small candidate sets only.

\subsection{Late Fusion Techniques\label{app:fusion_techniques}}
A late fusion function $f(q,a,\mathcal{M}): \mathcal{V}^{|q|} \times \mathcal{V}^{|a|} \times \mathcal{M} \rightarrow \mathbb{R}_+$ computes a relevance score between query $q$ and article $a$ by combining the ranked lists of articles $\mathcal{R}_m \subset \mathcal{C}$ returned separately by a set of retrieval models $\mathcal{M}$.

%in which an article ranked at position $i$ in a list of $N$ candidates is awarded $N-i+1$ points. The cumulative score for each article, aggregated across all ranked lists, determines the final ranking.
\paragraph{Borda count fusion (BCF)} uses a straightforward approach -- originally developed as a voting mechanism \citep{borda1781memoire} -- which combines the ranks from different systems linearly \citep{ho1994decision} such that
\begin{equation}
    f_{\textsc{bcf}}(q,a,\mathcal{M})=\sum_{m \in \mathcal{M}} |\mathcal{R}_m| - \pi_m(q,a) + 1,
\end{equation}
where $\pi_m(q,a) \in \left[1, |\mathcal{R}_m|\right]$ denotes the rank of article $a$ in the list of results returned by model $m$ for query $q$, i.e.,
\begin{equation}
    \pi_m(q, a) = 1+\sum_{a_i \in \mathcal{C}} \mathbbm{1}[s_m(q,a_i)>s_m(q, a)].
\end{equation}

%RRF is a rank-based method that assigns points to an article based on the inverse of its rank incremented by a tunable constant parameter. The points for each article are then summed across all ranked lists to derive the final ranking.
\paragraph{Reciprocal rank fusion (RRF)} refines the previous approach by introducing a non-linear weighting scheme that gives more emphasis to top-ranked documents \citep{cormack2009reciprocal}, i.e.,
\begin{equation}
    f_{\textsc{rrf}}(q,a,\mathcal{M})=\sum_{m \in \mathcal{M}} \frac{1}{k+\pi_m(q,a)},
\end{equation}
where $k > 0$ is a constant set to 60 by default.

%NSF scales the potentially unbounded system-specific scores before linearly combining them using predetermined weights. These weights can be equally distributed, as in CombSUM \citep{shaw1994combination}, or varied, with higher values effectively granting more importance to particular systems.
\paragraph{Normalized score fusion (NSF)} linearly combines the output relevance scores from distinct retrieval models \citep{lee1995combining} such that
\begin{equation}
    f_{\textsc{nsf}}(q,a,\mathcal{M})=\sum_{m \in \mathcal{M}} \alpha_m \hat{s}_m(q, a),
\end{equation}
where the scalars $\alpha_m$, controlling the relative importance of each model $m$ in the fused score, are non-negative and sum to one. These weights can be varied or uniformly distributed, as in CombSUM \citep{shaw1994combination}. Given that the original model-specific scores can be unbounded, they are generally normalized prior to fusion, using either min-max scaling where
\begin{equation}
    \hat{s}_m(q, a) = \frac{s_m(q, a) - \min_{i=1}^{|\mathcal{C}|} s_m(q, a_i)}{\max_{i=1}^{|\mathcal{C}|} s_m(q, a_i) - \min_{i=1}^{|\mathcal{C}|} s_m(q, a_i)},
\end{equation}
or z-score scaling such that
\begin{equation}
    \hat{s}_m(q, a) = \frac{s_m(q, a) - \mu_m(q)}{\sigma_m(q)},
\end{equation}
where $\mu_m(q)$ is the mean score across all candidate articles in the ranked list for query $q$ returned by model $m$, and $\sigma_m(q)$ denotes the standard deviation of these scores. Beyond these conventional scaling methods, we also investigate a percentile-based normalization, the rationale and specifics of which are elaborated in \Cref{sec:zeroshot_eval}.

\begin{table*}[t]
\centering
\resizebox{\textwidth}{!}{%
\begin{tabular}{l|rccc|ccc} 
\toprule
\textbf{French PLM Backbone} & \textbf{\#Params} & \textbf{Architecture} & \textbf{\#L} & \textbf{Pre-training} & \textbf{MRR\at10} & \textbf{R\at100} & \textbf{R\at500} \\
\midrule
DistilCamemBERT \citep{delestre2022distil} & 68.1M & BERT & 6 & \textsc{mlm}+\textsc{kl}+\textsc{cos} & \underline{0.268} & \underline{0.764} & \underline{0.879} \\ 
ELECTRA-fr{\sub{base}} \citep{schweter2020europeana} & 110.0M & BERT & 12 & \textsc{rtd} & 0.234 & 0.690 & 0.816 \\
CamemBERT{\sub{base}} \citep{martin2020camembert} & 110.6M & BERT & 12 & \textsc{mlm} & \textbf{0.285} & \textbf{0.778} & \textbf{0.891} \\
CamemBERTa{\sub{base}} \citep{antoun2023data} & 111.8M & DeBERTa & 12 & \textsc{rtd} & 0.248  & 0.696 & 0.822 \\ 
\bottomrule
\end{tabular}
}
\caption{In-domain retrieval performances on mMARCO-fr small dev set \citep{bonifacio2021mmarco} for single-vector dense representation models trained using various French pretrained autoencoding language models as their text embedding backbone. \textsc{mlm}, \textsc{rtd}, \textsc{kl}, and \textsc{cos} denote the masked language modeling \citep{devlin2019bert}, replaced token detection \citep{clark2020electra}, Kullback-Leibler divergence \citep{radford2018improving}, and negative cosine embedding \citep{sanh2019distil} training objectives, respectively. \#L indicates the number of encoder layers.}
\label{tab:french_backbone}
\end{table*}
%The best results are marked in \textbf{bold}, and the second best are \underline{underlined}.

\subsection{Evaluation Metrics\label{app:evaluation_metrics}}
Let $\operatorname{rel}(q,a): \mathcal{V}^m \times \mathcal{V}^n \rightarrow \{0,1\}$ be a binary relevance function, indicating whether an article $a$ from the corpus $\mathcal{C}$ is relevant to a query $q$. Assume that $\mathcal{R}_q = \{(i, a)\}_{i=1}^{k}$ denotes the ranked list of articles returned by a retrieval system, truncated at the top-$k$ results. We define the metrics mentioned in \Cref{subsec:experimental_setup} as follows.

\paragraph{Recall\at$k$.}
The metric quantifies the proportion of relevant articles retrieved within the top-$k$ ranked results for query $q$, compared to the total number of relevant articles in the corpus $\mathcal{C}$, i.e.,
\begin{equation}
    \text{R\at}k(q, \mathcal{R}_q) = \frac{\sum_{(i,a) \in \mathcal{R}_q} \operatorname{rel}(q,a)}{\sum_{a \in \mathcal{C}} \operatorname{rel}(q,a)}.
\end{equation}

\paragraph{Reciprocal rank\at$k$.}
The metric takes the inverse of the position at which the first relevant article appears within the top-$k$ results for query $q$, i.e.,
\begin{equation}
    \text{RR\at}k(q, \mathcal{R}_q) = \max _{(i,a) \in \mathcal{R}_q} \frac{\operatorname{rel}(q,a)}{i}.
\end{equation}

\paragraph{R-precision.}
The metric computes the ratio of relevant articles within the top-$N$ retrieved results for query $q$, where $N$ represents the total number of relevant articles for that query, i.e.,
\begin{equation}
    \text{RP}(q,\mathcal{R}_q) = \frac{\sum_{(i,a) \in \{\mathcal{R}_q\}_{i=1}^{N}} \operatorname{rel}(q,a)}{N}.
\end{equation}
For all metrics, we report the average scores over a set of $Q$ queries.

\subsection{Counting FLOPs \label{app:counting_flops}}
Below, we detail our methodology to estimate the inference complexity per query in terms of floating point operations (FLOPs). Except BM25, the main computational cost derives from the Transformer encoder's forward pass, executed once with bi-encoder models to encode the query, and repeatedly in cross-encoders to process each query-article pair. We leverage DeepSpeed's profiler to measure the forward pass cost of each neural retriever.\footnote{\url{https://www.deepspeed.ai/tutorials/flops-profiler/}} Queries are assumed to be 15 tokens and articles 157 tokens, as per their respective average lengths in LLeQA.

\paragraph{BM25.}
In the BM25 scoring formula, outlined in \Cref{eq:score_bm25}, several elements can be pre-computed and cached to streamline computations during inference. These include the inverse document frequency (IDF) for each term, the normalized document lengths adjusted by the parameters $k1$ and $b$, and the constant $(k1\!+\!1)$. For each query term and candidate document, the process involves four primary operations. First, the term frequency (TF), retrieved via a simple lookup, is multiplied by the pre-computed IDF and $(k1\!+\!1)$. The result is then added to the stored normalized document length. Finally, this sum is used as the denominator in dividing the product of the TF, IDF, and $(k1\!+\!1)$. These four operations -- two multiplications, one addition, and one division -- per term-document pair lead to an overall computational cost of $4\overbar{|q|}\mathcal{|C|}$ FLOPs for searching across the whole corpus.

\paragraph{SPLADE{\sub{fr-base}}.}
At indexing time, this model creates a pseudo-TF for each token $t$ in the vocabulary by scaling and rounding the corresponding activation weights in sparse article representations. This enables the construction of a pseudo text collection where each term $t$ is repeated $\text{TF}(t, a)$ times for article $a$. During inference, obtaining the query representation requires a single forward pass. For each non-zero term in that representation, the search process involves three core steps: accessing the inverted list for the term (a negligible lookup operation), multiplying the query term weight by each article term weight from that list, and adding each result to the corresponding article's score accumulator. Consequently, for each term-article pair, the operations include one multiplication and one addition. With $C_{\textsc{fw}}$ representing the cost of the encoder's forward pass, $\overbar{|\mathbf{h}^+_q|}$ the average number of non-zero terms in the query representation, and $\overbar{|\mathcal{L}|}$ the average length of the inverted lists for these terms, the total computational complexity is estimated as $C_{\textsc{fw}} + 2\overbar{|\mathbf{h}^+_q|}\overbar{|\mathcal{L}|}$ FLOPs.\footnote{On LLeQA, the model activates an average of 178 tokens per query, and the associated index features inverted lists of 378 elements on average.}

\paragraph{Single-vector dense bi-encoders.}
With these models, a brute-force search across all articles from corpus $\mathcal{C}$ necessitates $|\mathcal{C}|$ inner products between $d$-dimensional article representations -- each involving $d$ multiplications and $d\!-\!1$ additions. Consequently, the total inference cost amounts to $C_{\textsc{fw}} + (2d\!-\!1) |\mathcal{C}|$ operations.

\paragraph{ColBERT{\sub{fr-base}}.}
For each candidate article, this model computes \Cref{eq:score_colbert} with the query and candidate token representations of $d$ dimensions. For each query term, this computation requires $2d\overbar{|q|}\overbar{|a|}$ operations for token-level inner products, $\overbar{|q|}\overbar{|a|}$ to identify the row-wise max, and $\overbar{|q|}$ for the final average. When performing brute-force search across the entire corpus, the inference complexity is estimated as $C_{\textsc{fw}} + \overbar{|q|}^2 (2d\overbar{|a|} \!+ \!\overbar{|a|} \!+ \!1)|\mathcal{C}|$ FLOPs.
%The PLAID implementation of ColBERT contains complex engineering schemes and low-level optimization such as centroid interaction and fast kernels.

\paragraph{monoBERT{\sub{fr-base}}.}
This model requires one forward pass per article to assess, incurring a high computational cost that typically limits their use to re-ranking a set of candidates returned by a cheaper retrieval model. To reflect that practice, we report the number of operations needed to score a fixed set of 1000 articles, resulting in $10^3C_{\textsc{fw}}$ FLOPs.

%inference compute cost per query using exact search
%Estimation for IVF: https://arxiv.org/pdf/2305.19435 (Appendix B1)
%Estimation for ColBERT: https://arxiv.org/pdf/2304.01982 (Appendix B)
%Estimation forward pass of Transformer: https://arxiv.org/pdf/2001.08361 (Table 1)

\begin{table*}[t]
\centering
\resizebox{\textwidth}{!}{%
\begin{tabular}{ll|cccc|cccc}
\toprule
& \multicolumn{1}{c}{\textbf{Training data} $(\rightarrow)$} & \multicolumn{4}{c}{\textbf{mMARCO-fr}} & \multicolumn{4}{c}{\textbf{LLeQA}} \\ \cmidrule(r){2-2} \cmidrule(r){3-6} \cmidrule(r){7-10}
& \multicolumn{1}{c}{\textbf{Learned model} $(\rightarrow)$} & \textbf{DPR{\sub{fr-base}}} & \textbf{SPLADE{\sub{fr-base}}} & \textbf{ColBERT{\sub{fr-base}}} & \multicolumn{1}{c}{\textbf{monoBERT{\sub{fr-base}}}} & \textbf{DPR{\sub{fr-lex}}} & \textbf{SPLADE{\sub{fr-lex}}} & \textbf{ColBERT{\sub{fr-lex}}} & \textbf{monoBERT{\sub{fr-lex}}}\\
\midrule
\multicolumn{2}{l}{\textbf{Configuration}} & & \\
    & Max query length      & 128       & 32        & 32        & -         & 512       & 64        & 64        & -         \\
    & Max article length    & 128       & 128       & 128       & $512-|q|$ & 512       & 512       & 512       & $512-|q|$ \\
    & Pooling strategy      & mean      & max       & -         & cls       & mean      & max       & -         & cls       \\
    & Similarity function   & cos       & cos       & cos       & -         & cos       & cos       & cos       & -         \\
\midrule
\multicolumn{2}{l}{\textbf{Hyperparameters}} & & \\
    & Steps                 & 66k       & 100k      & 200k      & 20k       & 1k        & 2k        & 1k        & 2k        \\
    & Batch size            & 152       & 128       & 128       & 128       & 64        & 32        & 64        & 64        \\
    & Optimizer             & AdamW     & AdamW     & AdamW     & AdamW     & AdamW     & AdamW     & AdamW     & AdamW     \\
    & Weight decay          & 0.01      & 0.01      & 0.0       & 0.01      & 0.01      & 0.01      & 0.0       & 0.01      \\
    & Peak learning rate    & 2e-5      & 2e-5      & 5e-6      & 2e-5      & 2e-5      & 2e-5      & 5e-6      & 2e-5      \\
    & Learning rate decay   & linear    & linear    & linear    & constant  & constant  & constant  & constant  & constant  \\
    & Warm-up ratio         & 0.01      & 0.04      & 0.1       & 0.0       & 0.0       & 0.0       & 0.0       & 0.0       \\
    & Gradient clipping     & 1.0       & 1.0       & 1.0       & 1.0       & 1.0       & 1.0       & 1.0       & 1.0       \\
    & Softmax temperature   & 0.05      & 0.05      & 1.0       & -         & 0.05      & 0.05      & 1.0       & -         \\
\midrule
\multicolumn{2}{l}{\textbf{Energy}} & & \\
    & Hardware                      & V100  & H100  & H100  & H100  & H100  & H100  & H100  & H100  \\
    & Thermal design power (W)      & 300   & 310   & 310   & 310   & 310   & 310   & 310   & 310   \\
    & Training time (h)             & 14.1  & 12.9  & 18.4  & 1.5   & 0.22  & 0.30  & 0.18  & 0.17  \\
    & Power consumption (kWh)       & 4.2   & 4.0   & 5.7   & 0.5   & 0.07  & 0.09  & 0.06  & 0.05  \\
    & Carbon emission (kgCO$_2$eq)  & 1.8   & 1.7   & 2.5   & 0.2   & 0.03  & 0.04  & 0.03  & 0.02  \\
\bottomrule
\end{tabular}%
}
\caption{Implementation details for our learned domain-general (\textsc{fr-base}) and domain-specific (\textsc{fr-lex}) retrievers.}
\label{tab:implementation}
\end{table*}

\begin{figure*}[htb!]
\centering
\begin{subfigure}[t]{.32\textwidth}
  \centering
  \includegraphics[width=\linewidth]{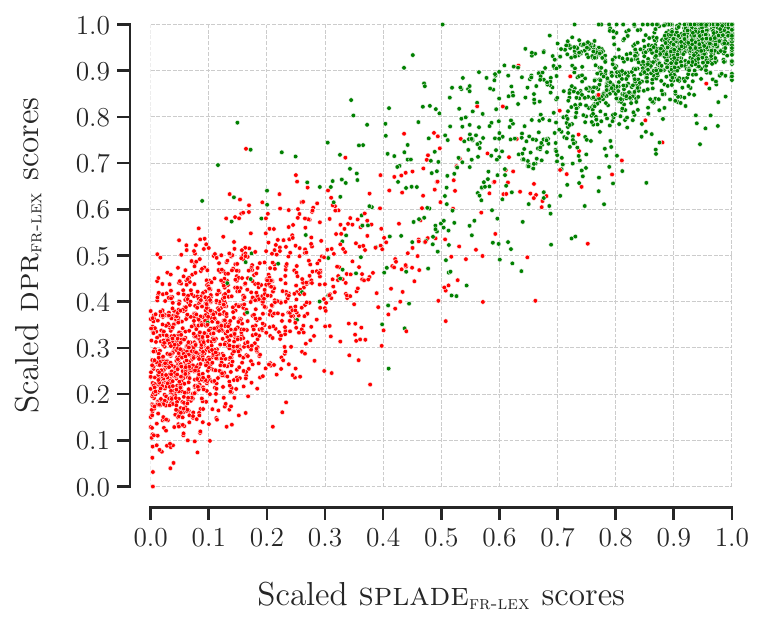}
\end{subfigure}
\begin{subfigure}[t]{.32\textwidth}
  \centering
  \includegraphics[width=\linewidth]{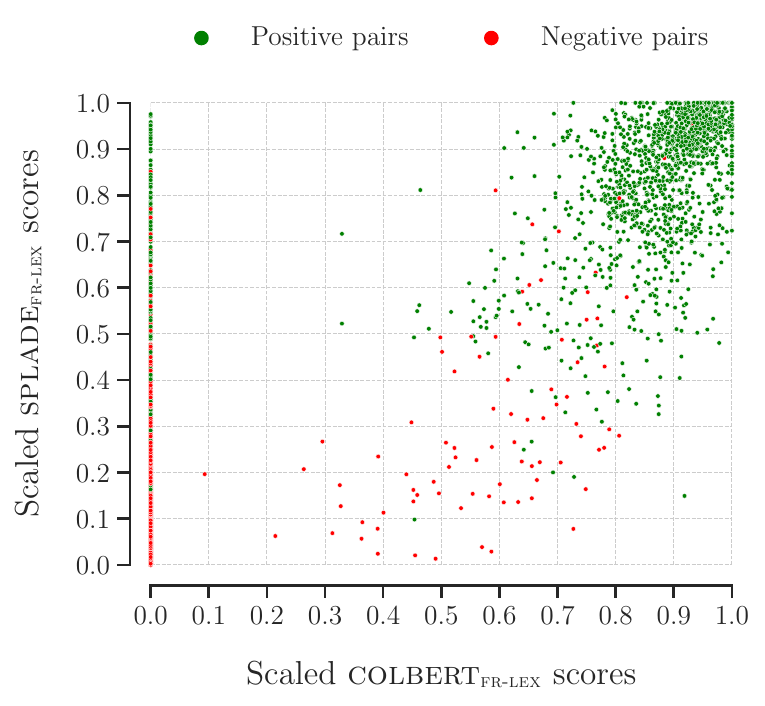}
\end{subfigure}
\begin{subfigure}[t]{.32\textwidth}
  \centering
  \includegraphics[width=\linewidth]{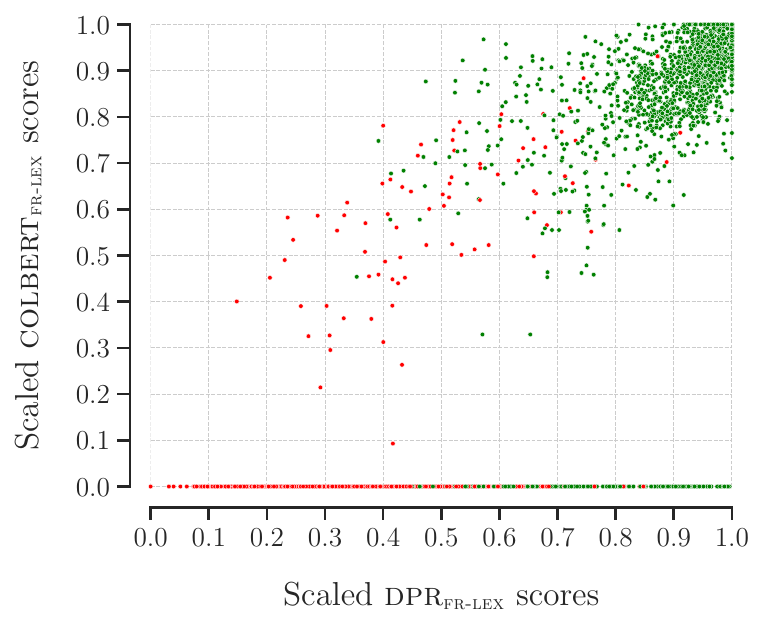}
\end{subfigure}
% \begin{subfigure}[t]{.32\textwidth}
%   \centering
%   \includegraphics[width=\linewidth]{img/score_relationship_minmax_bm25-dpr_indomain.pdf}
% \end{subfigure}
% \begin{subfigure}[t]{.32\textwidth}
%   \centering
%   \includegraphics[width=\linewidth]{img/score_relationship_minmax_bm25-colbert_indomain.pdf}
% \end{subfigure}
% \begin{subfigure}[t]{.32\textwidth}
%   \centering
%   \includegraphics[width=\linewidth]{img/score_relationship_minmax_bm25-splade_indomain.pdf}
% \end{subfigure}
\caption{Distribution of paired relevance scores from our learned specialized retrievers on around 3,000 query-article pairs from the LLeQA dev set, evenly balanced between positive and negative instances.}
\label{fig:indomain_complentarity}
\end{figure*}

\section{Implementation Details \label{app:implementation}}

\subsection{Embedding Backbone \label{app:backbone}}
To ensure a fair comparison between the different matching paradigms detailed in \Cref{subsec:retrieval_models}, irrespective of the underlying text embedding model's capacity, we choose to exploit the same pretrained autoencoding language model across all our neural retrievers. To explore the efficacy of existing French embedding models for text retrieval, we finetune four prominent pretrained models on mMARCO-fr, including CamemBERT{\sub{base}} \citep{martin2020camembert}, ELECTRA-fr{\sub{base}} \citep{schweter2020europeana}, DistilCamemBERT \citep{delestre2022distil}, and CamemBERTa{\sub{base}} \citep{antoun2023data}. We limit our investigation to the performance of single-vector dense bi-encoders to minimize environmental impact. \Cref{tab:french_backbone} presents the results on the mMARCO-fr small dev set, revealing that CamemBERT{\sub{base}} significantly outperforms the other French text encoders. Following these findings, we select this model as the common backbone encoder for all our neural retrievers.

\subsection{Optimization \label{app:optimization}}
\Cref{tab:implementation} provides details on our models' configuration, training hyperparameters, and energy consumption. Training and GPU-based experiments are conducted one a single 80GB NVIDIA H100, while CPU-based evaluations are performed on a server with a 64-core AMD EPYC 7763 CPU at 3.20GHz and 500GB of RAM. We implement, train, tune, and monitor our models using the following Python libraries: \href{https://github.com/pytorch/pytorch}{\texttt{pytorch}} \citep{paszke2019pytorch}, \href{https://github.com/huggingface/transformers}{\texttt{transformers}} \citep{wolf2020transformers}, \href{https://github.com/UKPLab/sentence-transformers}{\texttt{sentence-transformers}} \citep{reimers2019sentence}, \href{https://github.com/stanford-futuredata/ColBERT}{\texttt{colbert-ai}} \citep{khattab2020colbert}, and \href{https://github.com/wandb/wandb}{\texttt{wandb}} \citep{biewald2020wandb}.

\begin{table*}[t]
  \begin{minipage}[t]{0.48\textwidth}
    \vspace{0pt}
    \centering
    \resizebox{\columnwidth}{!}{%
    \begin{tabular}{l|cccc}
    \toprule
    \textbf{\#} & \textbf{BM25} & \textbf{DPR{\sub{fr-base}}} & \textbf{SPLADE{\sub{fr-base}}} & \textbf{ColBERT{\sub{fr-base}}} \\
    \midrule
    \multicolumn{5}{c}{\textit{Min-max scaling}} \\
    \shade{5}  & .50 & 0   & .50 & 0   \\
    \shade{6}  & 0   & .25 & 0   & .75 \\
    \shade{7}  & .40 & .60 & 0   & 0   \\
    \shade{8}  & .40 & 0   & 0   & .60 \\
    \shade{9}  & 0   & .70 & .30 & 0   \\
    \shade{10} & 0   & 0   & .20 & .80 \\
    \shade{11} & .25 & .25 & .50 & 0   \\
    \shade{12} & .35 & 0   & .40 & .25 \\
    \shade{13} & .35 & .25 & 0   & .40 \\
    \shade{14} & 0   & .10 & .20 & .70 \\
    \shade{15} & .30 & .35 & .10 & .25 \\
    \midrule
    \multicolumn{5}{c}{\textit{Z-score scaling}} \\
    \shade{5}  & .40 & 0   & .60 & 0   \\
    \shade{6}  & 0   & .25 & 0   & .75 \\
    \shade{7}  & .30 & .70 & 0   & 0   \\
    \shade{8}  & .25 & 0   & 0   & .75 \\
    \shade{9}  & 0   & .80 & .20 & 0   \\
    \shade{10} & 0   & 0   & .20 & .80 \\
    \shade{11} & .20 & .40 & .40 & 0   \\
    \shade{12} & .20 & 0   & .40 & .40 \\
    \shade{13} & .20 & .30 & 0   & .50 \\
    \shade{14} & 0   & .40 & .10 & .50 \\
    \shade{15} & .15 & .45 & .10 & .30 \\
    \midrule
    \multicolumn{5}{c}{\textit{Percentile scaling}} \\
    \shade{5}  & .60 & 0   & .40 & 0   \\
    \shade{6}  & 0   & .05 & 0   & .95 \\
    \shade{7}  & .50 & .50 & 0   & 0   \\
    \shade{8}  & .40 & 0   & 0   & .60 \\
    \shade{9}  & 0   & .85 & .15 & 0   \\
    \shade{10} & 0   & 0   & .20 & .80 \\
    \shade{11} & .45 & .05 & .50 & 0   \\
    \shade{12} & .55 & 0   & .35 & .10 \\
    \shade{13} & .50 & .40 & 0   & .10 \\
    \shade{14} & 0   & .05 & .70 & .25 \\
    \shade{15} & .50 & .05 & .40 & .05 \\
    \bottomrule
    \end{tabular}%
    }
    \caption{Optimally tuned weights for the normalized score fusion results presented in \Cref{tab:outdomain_fusion} (zero-shot).}
    \label{tab:tuned_weights_outdomain}
  \end{minipage}
  \hfill
  \begin{minipage}[t]{0.4655\textwidth}
    \vspace{0pt}
    \centering
    \resizebox{\columnwidth}{!}{%
    \begin{tabular}{l|cccc}
    \toprule
    \textbf{\#} & \textbf{BM25} & \textbf{DPR{\sub{fr-lex}}} & \textbf{SPLADE{\sub{fr-lex}}} & \textbf{ColBERT{\sub{fr-lex}}} \\
    \midrule
    \multicolumn{5}{c}{\textit{Min-max scaling}} \\
    \shade{5}  & .15 & 0   & .85 & 0   \\
    \shade{6}  & 0   & .85 & 0   & .15 \\
    \shade{7}  & .10 & .90 & 0   & 0   \\
    \shade{8}  & .15 & 0   & 0   & .85 \\
    \shade{9}  & 0   & .70 & .30 & 0   \\
    \shade{10} & 0   & 0   & .85 & .15 \\
    \shade{11} & .05 & .60 & .35 & 0   \\
    \shade{12} & .15 & 0   & .75 & .10 \\
    \shade{13} & .10 & .80 & 0   & .10 \\
    \shade{14} & 0   & .60 & .25 & .15 \\
    \shade{15} & .05 & .60 & .30 & .05 \\
    \midrule
    \multicolumn{5}{c}{\textit{Z-score scaling}} \\
    \shade{5}  & .10 & 0   & .90 & 0   \\
    \shade{6}  & 0   & .65 & 0   & .35 \\
    \shade{7}  & .05 & .95 & 0   & 0   \\
    \shade{8}  & .05 & 0   & 0   & .95 \\
    \shade{9}  & 0   & .70 & .30 & 0   \\
    \shade{10} & 0   & 0   & .75 & .25 \\
    \shade{11} & .05 & .55 & .40 & 0   \\
    \shade{12} & .05 & 0   & .75 & .20 \\
    \shade{13} & .05 & .80 & 0   & .15 \\
    \shade{14} & 0   & .60 & .25 & .15 \\
    \shade{15} & .05 & .80 & .05 & .10 \\
    \midrule
    \multicolumn{5}{c}{\textit{Percentile scaling}} \\
    \shade{5}  & .05 & 0   & .95 & 0   \\
    \shade{6}  & 0   & .95 & 0   & .05 \\
    \shade{7}  & .05 & .95 & 0   & 0   \\
    \shade{8}  & .10 & 0   & 0   & .90 \\
    \shade{9}  & 0   & .85 & .15 & 0   \\
    \shade{10} & 0   & 0   & .95 & .05 \\
    \shade{11} & .05 & .45 & .50 & 0   \\
    \shade{12} & .05 & 0   & .90 & .05 \\
    \shade{13} & .05 & .75 & 0   & .20 \\
    \shade{14} & 0   & .85 & .10 & .05 \\
    \shade{15} & .05 & .40 & .50 & .05 \\
    \bottomrule
    \end{tabular}%
    }
    \caption{Optimally tuned weights for the normalized score fusion results presented in \Cref{tab:indomain_fusion} (in-domain).}
    \label{tab:tuned_weights_indomain}
  \end{minipage}
\end{table*}

\section{Additional Results\label{app:additional_results}}

\subsection{Complementarity of Specialized Models\label{app:indomain_complentarity}}
To understand why fusion does not enhance the performance of specialized retrievers, we examine the complementarity of their relevance signals in \Cref{fig:indomain_complentarity}. We sample approximately 1,500 positive query-article pairs from the LLeQA dev set, along with an equal number of random negatives, and gather the scores assigned by the different models to each pair. Contrary to the zero-shot context, we find that the output scores from the specialized models align closely, as shown by the linear distribution of paired scores in \Cref{fig:indomain_complentarity}. Pairs that receive high relevance scores from one system typically receive similar scores from others, and the same applies for lower scores. We hypothesize that since all retrieval models were trained on the exact same domain-specific dataset using the same primary contrastive learning objective, they converged towards learning similar relevance signals, with some models like DPR{\sub{fr-lex}} developing more nuanced ones. Consequently, fusing models that have learned related signals, but with varying levels of accuracy, generally results in degraded performance compared to using the best model alone.

\subsection{Weight Tuning in NSF}
\begin{figure*}[t]
\centering
\begin{subfigure}[t]{.33\textwidth}
  \centering
  \includegraphics[width=\linewidth]{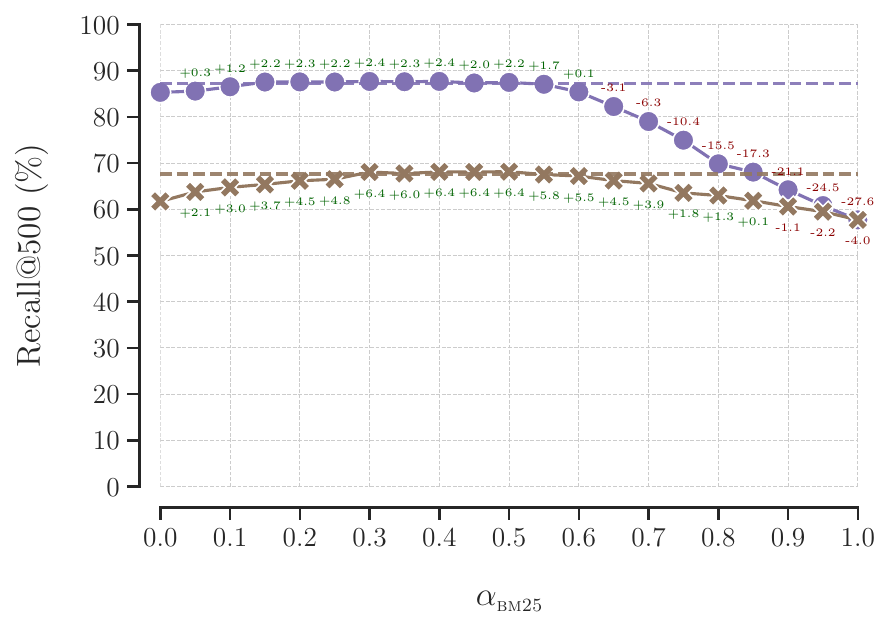}
\end{subfigure}
\begin{subfigure}[t]{.33\textwidth}
  \centering
  \includegraphics[width=\linewidth]{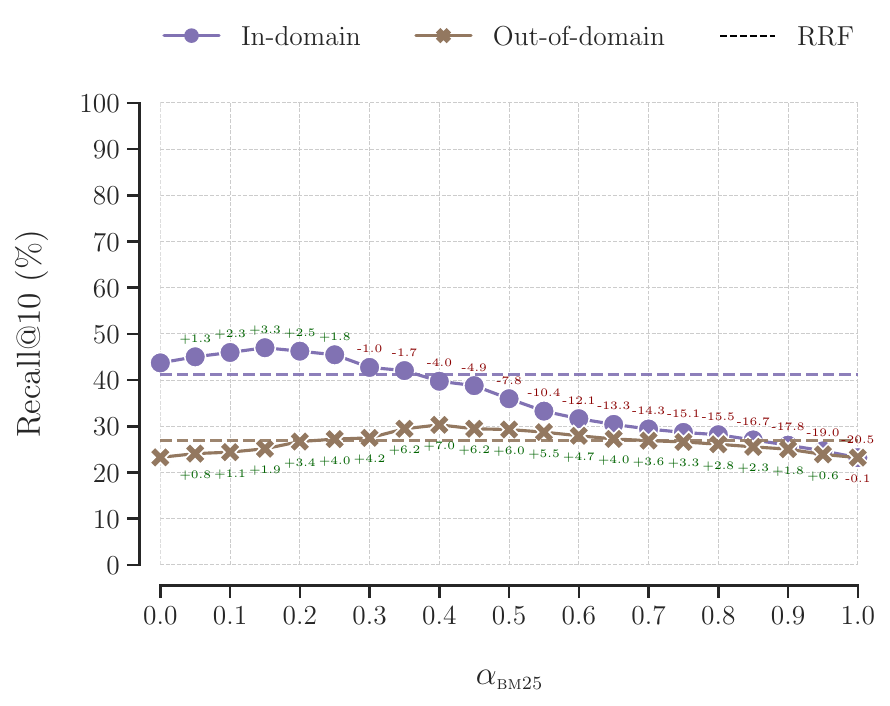}
\end{subfigure}
\begin{subfigure}[t]{.329\textwidth}
  \centering
  \includegraphics[width=\linewidth]{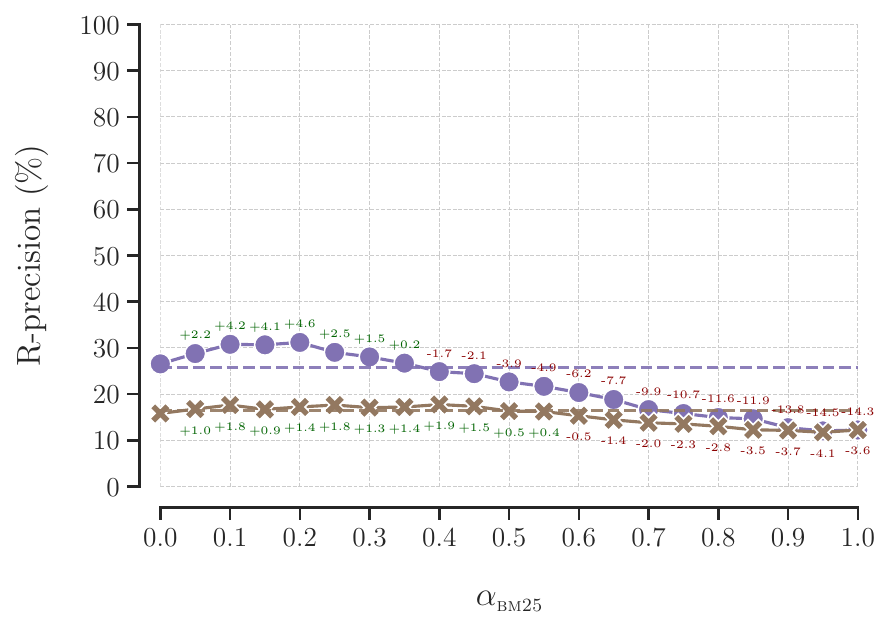}
\end{subfigure}
\caption{Effect of weight tuning in NSF between BM25 \& ColBERT{\sub{fr-\{\textcolor[HTML]{8172b3}{\textbf{lex}},\textcolor[HTML]{937860}{\textbf{base}}\}}} on LLeQA dev set.}
\label{fig:effect_weight_tuning_bm25-colbert}
\end{figure*}

\begin{figure*}[t]
\centering
\begin{subfigure}[t]{.33\textwidth}
  \centering
  \includegraphics[width=\linewidth]{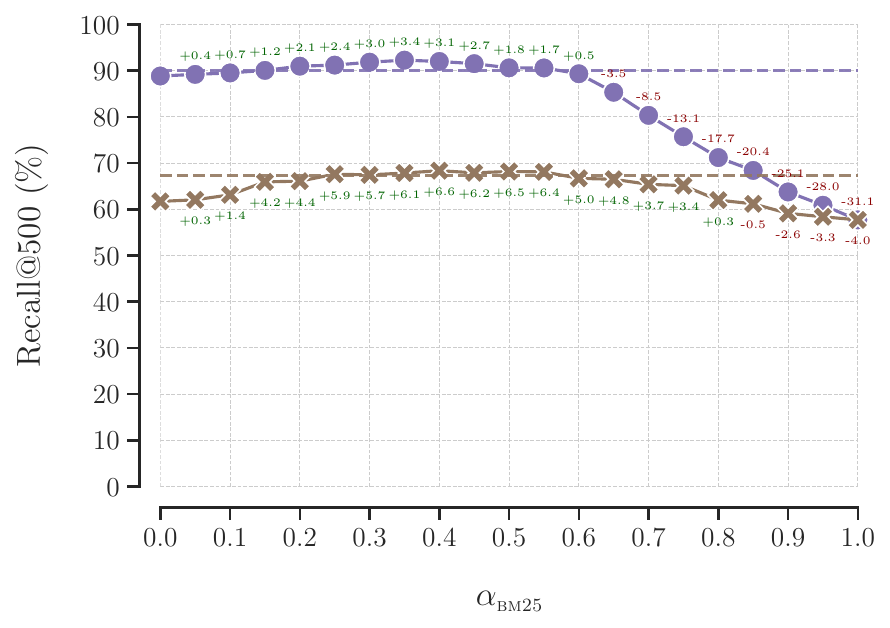}
\end{subfigure}
\begin{subfigure}[t]{.33\textwidth}
  \centering
  \includegraphics[width=\linewidth]{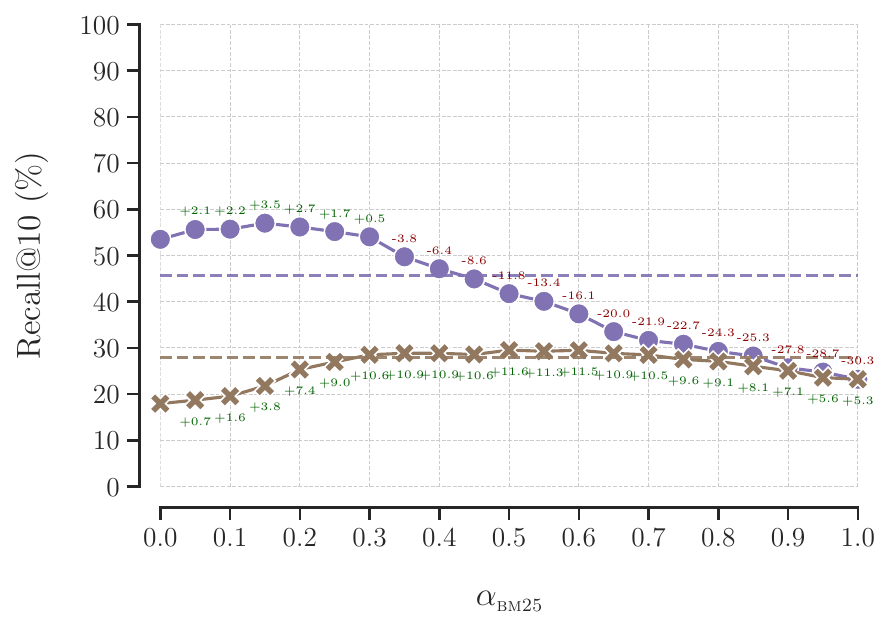}
\end{subfigure}
\begin{subfigure}[t]{.329\textwidth}
  \centering
  \includegraphics[width=\linewidth]{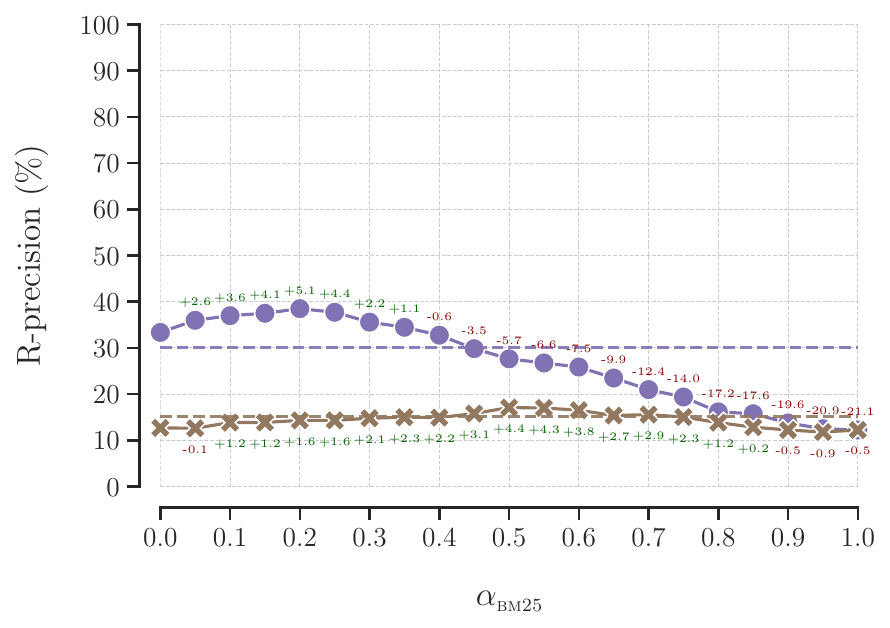}
\end{subfigure}
\caption{Effect of weight tuning in NSF between BM25 \& SPLADE{\sub{fr-\{\textcolor[HTML]{8172b3}{\textbf{lex}},\textcolor[HTML]{937860}{\textbf{base}}\}}} on LLeQA dev set.}
\label{fig:effect_weight_tuning_bm25-splade}
\end{figure*}

\begin{figure*}[t]
\centering
\begin{subfigure}[t]{.33\textwidth}
  \centering
  \includegraphics[width=\linewidth]{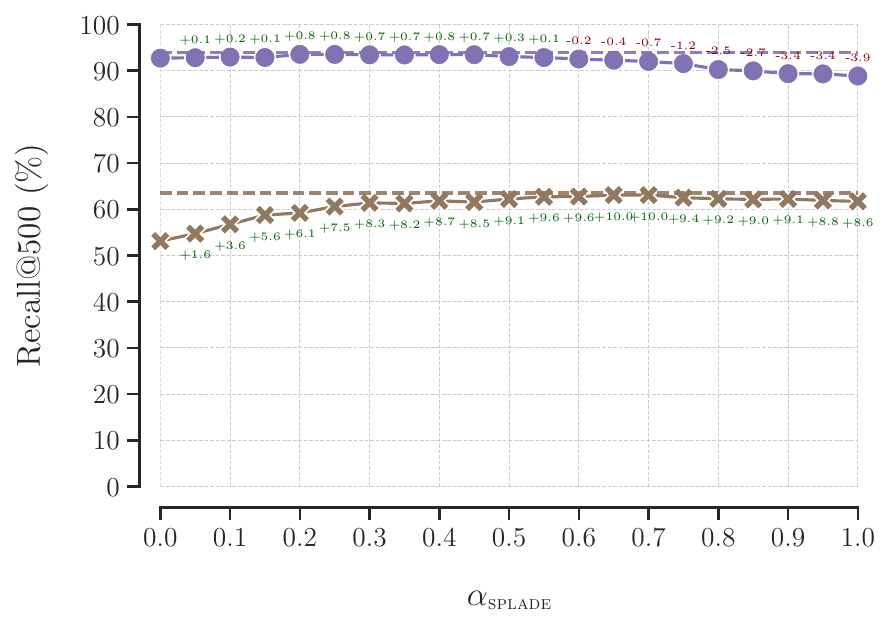}
\end{subfigure}
\begin{subfigure}[t]{.33\textwidth}
  \centering
  \includegraphics[width=\linewidth]{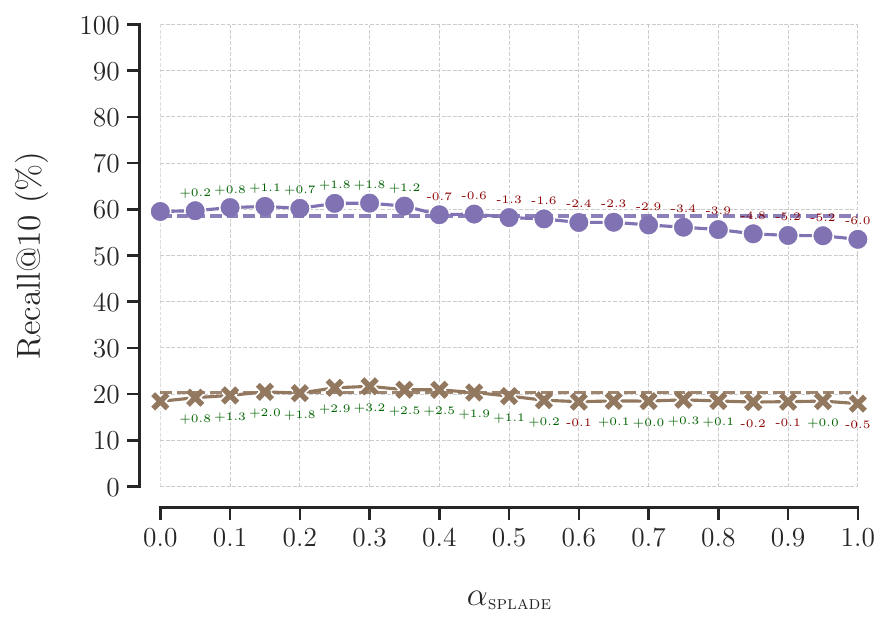}
\end{subfigure}
\begin{subfigure}[t]{.329\textwidth}
  \centering
  \includegraphics[width=\linewidth]{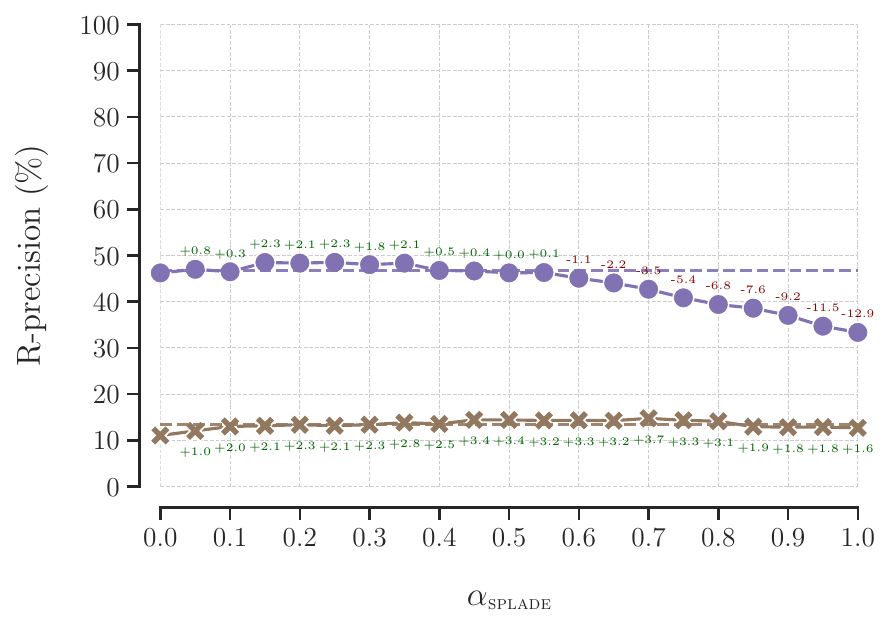}
\end{subfigure}
\caption{Effect of weight tuning in NSF between SPLADE{\sub{fr-\{\textcolor[HTML]{8172b3}{\textbf{lex}},\textcolor[HTML]{937860}{\textbf{base}}\}}} \& DPR{\sub{fr-\{\textcolor[HTML]{8172b3}{\textbf{lex}},\textcolor[HTML]{937860}{\textbf{base}}\}}} on LLeQA dev set.}
\label{fig:effect_weight_tuning_splade-dpr}
\end{figure*}

\begin{figure*}[t]
\centering
\begin{subfigure}[t]{.33\textwidth}
  \centering
  \includegraphics[width=\linewidth]{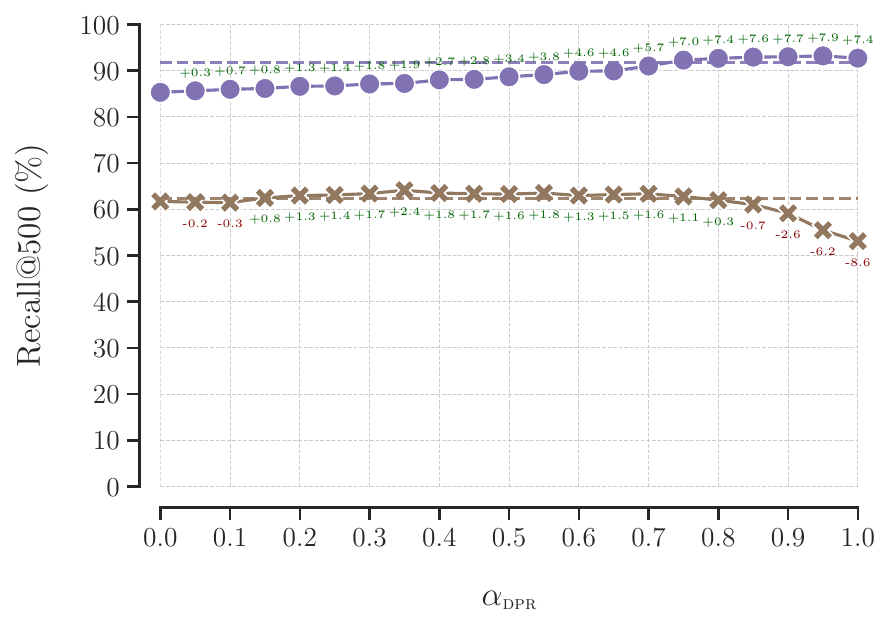}
\end{subfigure}
\begin{subfigure}[t]{.33\textwidth}
  \centering
  \includegraphics[width=\linewidth]{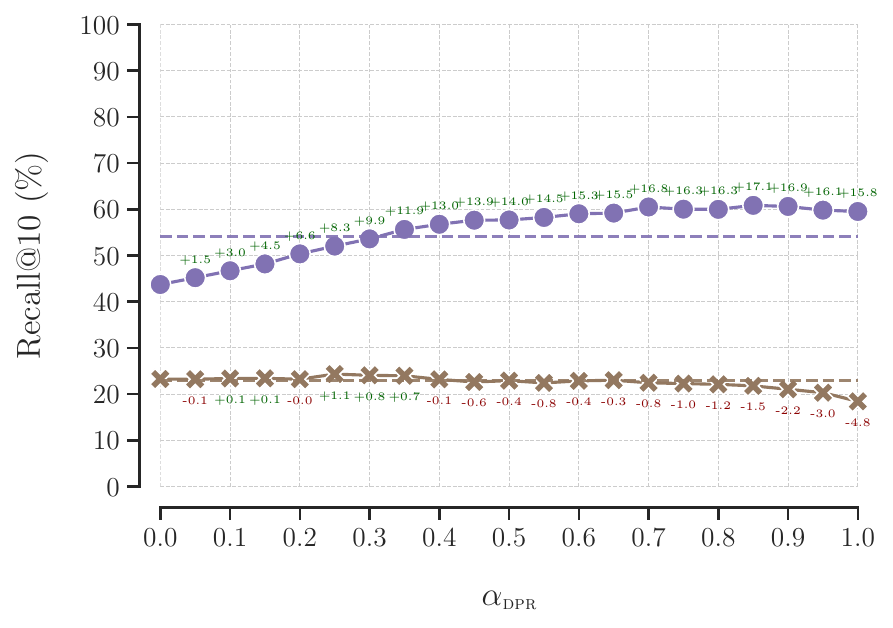}
\end{subfigure}
\begin{subfigure}[t]{.329\textwidth}
  \centering
  \includegraphics[width=\linewidth]{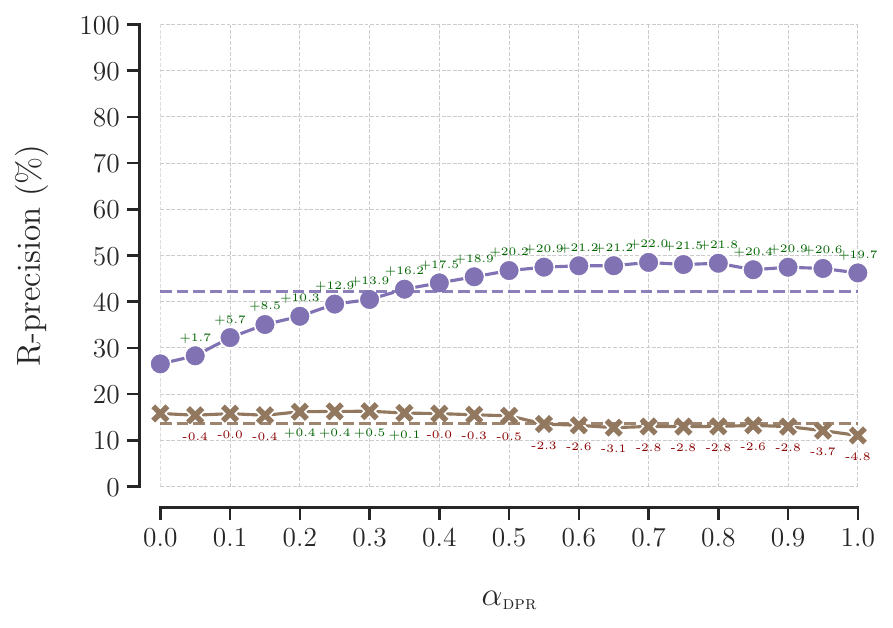}
\end{subfigure}
\caption{Effect of weight tuning in NSF between DPR{\sub{fr-\{\textcolor[HTML]{8172b3}{\textbf{lex}},\textcolor[HTML]{937860}{\textbf{base}}\}}} \& ColBERT{\sub{fr-\{\textcolor[HTML]{8172b3}{\textbf{lex}},\textcolor[HTML]{937860}{\textbf{base}}\}}} on LLeQA dev set.}
\label{fig:effect_weight_tuning_dpr-colbert}
\end{figure*}

\begin{figure*}[t]
\centering
\begin{subfigure}[t]{.33\textwidth}
  \centering
  \includegraphics[width=\linewidth]{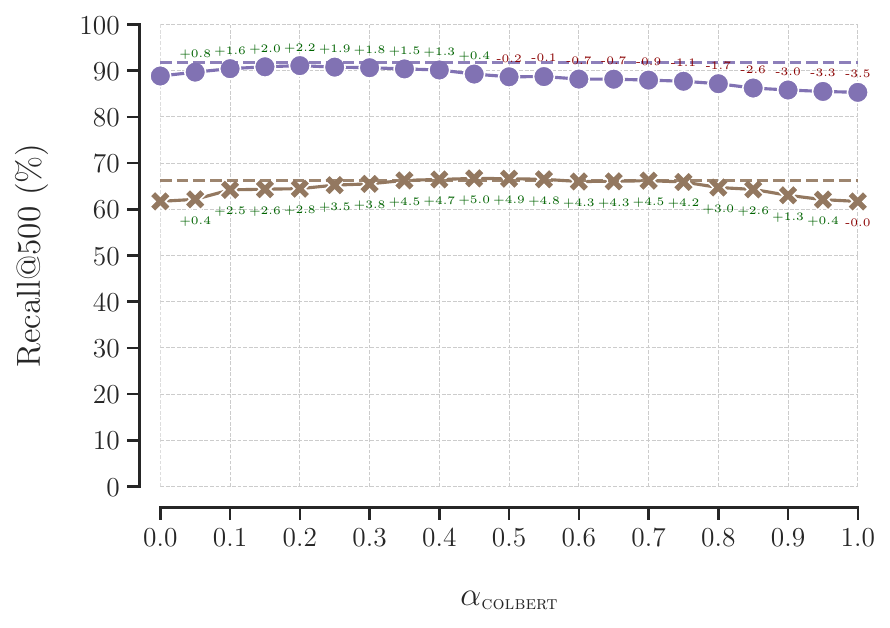}
\end{subfigure}
\begin{subfigure}[t]{.33\textwidth}
  \centering
  \includegraphics[width=\linewidth]{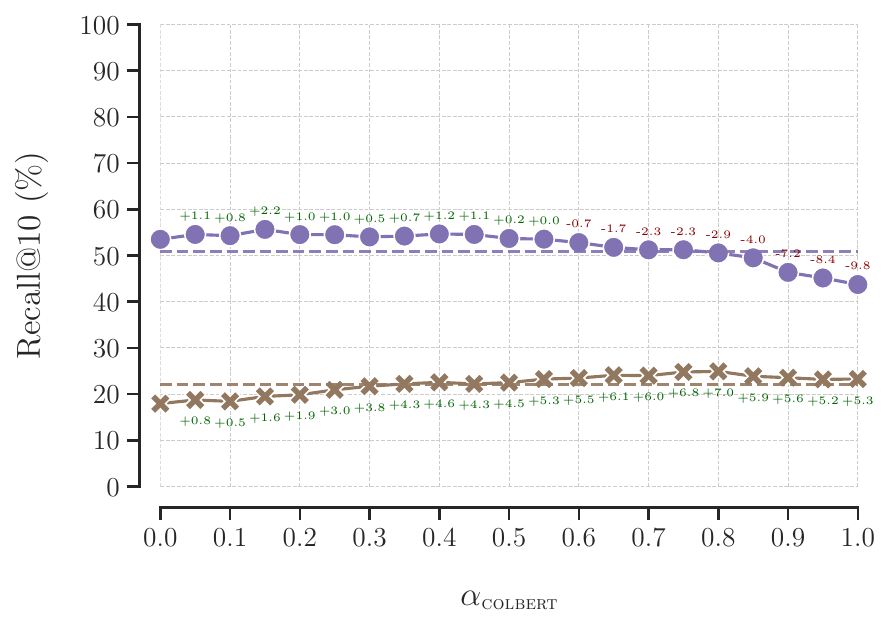}
\end{subfigure}
\begin{subfigure}[t]{.329\textwidth}
  \centering
  \includegraphics[width=\linewidth]{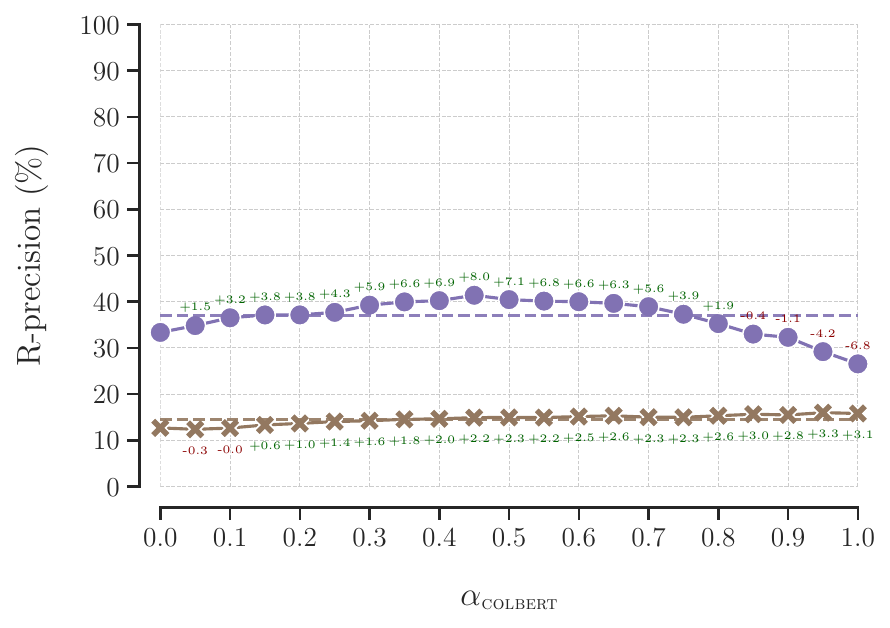}
\end{subfigure}
\caption{Effect of weight tuning in NSF between ColBERT{\sub{fr-\{\textcolor[HTML]{8172b3}{\textbf{lex}},\textcolor[HTML]{937860}{\textbf{base}}\}}} \& SPLADE{\sub{fr-\{\textcolor[HTML]{8172b3}{\textbf{lex}},\textcolor[HTML]{937860}{\textbf{base}}\}}} on LLeQA dev set.}
\label{fig:effect_weight_tuning_colbert-splade}
\end{figure*}

\Cref{tab:tuned_weights_outdomain} an \Cref{tab:tuned_weights_indomain} present the optimal weights assigned to each retrieval system in zero-shot and in-domain contexts, respectively, when using normalized score fusion (NSF). These weights were meticulously determined through extensive tuning on the LLeQA dev set. Additionally, Figures \labelcref{fig:effect_weight_tuning_bm25-colbert} to \labelcref{fig:effect_weight_tuning_colbert-splade} illustrate the variation in performance based on the weights assigned to pairs of retrieval systems.

\end{document}